\documentclass{article}

\PassOptionsToPackage{numbers, compress}{natbib}

\usepackage[preprint]{neurips_2024}

\usepackage[utf8]{inputenc} 
\usepackage[T1]{fontenc}    
\usepackage{hyperref}       
\usepackage{url}            
\usepackage{booktabs}       
\usepackage{amsfonts}       
\usepackage{nicefrac}       
\usepackage{microtype}      
\usepackage{xcolor}         
\usepackage{graphicx}
\usepackage{amsmath}
\usepackage{multirow}
\usepackage{colortbl}
\usepackage{wrapfig}
\usepackage{subcaption}
\usepackage{algorithm}
\usepackage{algpseudocode}
\usepackage{pifont}
\usepackage{titlesec}

\definecolor{lightergray}{rgb}{0.9,0.9,0.9}
\definecolor{darkgreen}{rgb}{0., 0.5, 0.}
\newcommand{\prevdiffusion}{\epsilon_{\theta_{i-1}} (\mathbf{x}_t, t, y)}
\newcommand{\currdiffusion}{\epsilon_{\theta_{i}} (\mathbf{x}_t, t, y)}
\newcommand{\currclassifier}{f_{\phi_i} (y|\mathbf{x})}
\newcommand{\prevclassifier}{f_{\phi_{i-1}} (y|\mathbf{x})}
\newcommand{\ours}{GUIDE}

\newcommand{\cmark}{\ding{51}}%
\newcommand{\xmark}{\ding{55}}

\title{GUIDE: Guidance-based Incremental Learning with Diffusion Models}

\author{%
    Bartosz Cywiński \\
  Warsaw University of Technology \\
  \texttt{bartosz.cywinski.stud@pw.edu.pl} \\
  \And
  Kamil Deja \\
  IDEAS NCBR \\
  Warsaw University of Technology \\
  \texttt{kamil.deja@pw.edu.pl} \\
  \And
  Tomasz Trzciński \\
  IDEAS NCBR \\
  Warsaw University of Technology \\
  Tooploox \\
  \texttt{tomasz.trzcinski@ideas-ncbr.pl} \\
  \And
  Bartłomiej Twardowski \\
  IDEAS NCBR \\
  Computer Vision Center \\
  Universitat Aut\`{o}noma de Barcelona \\
  \texttt{bartlomiej.twardowski@ideas-ncbr.pl} \\
  \And
  Łukasz Kuciński \\
  IDEAS NCBR \\
  University of Warsaw \\
  Polish Academy of Sciences \\
  \texttt{lukasz.kucinski@ideas-ncbr.pl} \\
}

\begin{document}

\maketitle

\begin{abstract}
We introduce \ours{}, a novel continual learning approach that directs diffusion models to rehearse samples at risk of being forgotten. 
Existing generative strategies combat catastrophic forgetting by randomly sampling rehearsal examples from a generative model. Such an approach contradicts buffer-based approaches where sampling strategy plays an important role.
We propose to bridge this gap by incorporating classifier guidance into the diffusion process to produce rehearsal examples specifically targeting information forgotten by a continuously trained model. This approach enables the generation of samples from preceding task distributions, which are more likely to be misclassified in the context of recently encountered classes. 
Our experimental results show that \ours{} significantly reduces catastrophic forgetting,
outperforming conventional random sampling approaches and surpassing recent state-of-the-art methods in continual learning with generative replay.
\end{abstract}

\section{Introduction}
A typical machine learning pipeline involves training a model on a static dataset and deploying it in a task with a similar data distribution. This assumption frequently proves impractical in real-world scenarios, where models encounter a constantly evolving set of objectives. To address this issue, Continual Learning (CL) methods try to accumulate knowledge from separate tasks while overcoming limited knowledge transfer and catastrophic forgetting~\cite{1999french}. 
Replay-based CL strategies employ a memory buffer to retain examples from preceding tasks, which are then mixed with samples from the current task. In order for this approach to be effective, it requires a well-designed sampling strategy such as approximating the true data distribution with stored samples~\cite{rebuffi2017icarl,isele2018selective}, maximizing the diversity of samples in a buffer~\cite{zaeemzadeh2019iterative},
or hard negative mining~\cite{jin2021gradient}.

\begin{wrapfigure}{R}{0.5\textwidth}
\vskip -0.1in
\centering
\includegraphics[width=0.5\textwidth]{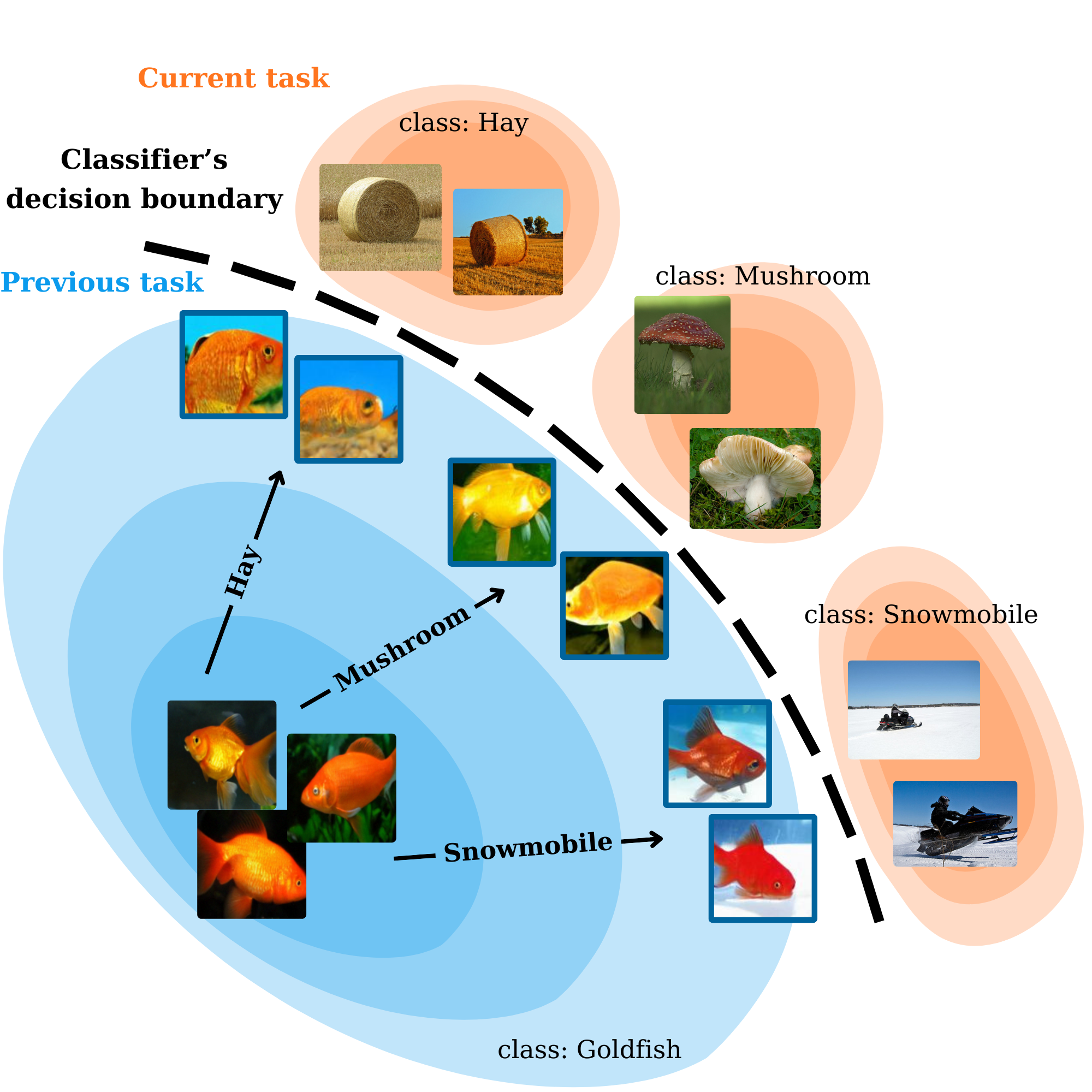}
\caption{Rehearsal sampling in \ours{}.
We guide the denoising process of a diffusion model trained on the previous task (\textcolor{blue}{blue}) toward classes from the current task (\textcolor{orange}{orange}).
The replay samples, highlighted with \textcolor{blue}{blue borders},
share features with the examples from the current task, which may be related to characteristics such as color or background (e.g., fishes on a snowy background when guided to \textit{snowmobile}).
Generative rehearsal on such samples positioned near the classifier's decision boundary successfully mitigates catastrophic forgetting.}

\label{fig:teaser}
\end{wrapfigure}

Despite offering decent performance, these methods are unsuitable for certain applications, e.g., due to limited scalability, memory constraints, or privacy restrictions. 
To mitigate this issue, \citet{shin2017continual} introduced Deep Generative Replay (DGR) that substitutes the memory buffer with a Generative Adversarial Networks (GAN)~\cite{2014goodfellow},
which was later extended with various generative models~\cite{van2020brain, scardapane2020pseudo,gao2023ddgr}. 
Although generative rehearsal approaches mitigate the limitations of buffer-based methods, relatively little attention was paid to understanding and controlling what samples they generate, simply relying on random generations. This contradicts the importance of sampling strategies proposed for buffer-based techniques.

In this work, we propose to bridge this gap and introduce \ours{}: a new continual learning method with Denoising Deep Probabilistic Models (DDPM) ~\cite{sohl2015deep,ho2020denoising}. We first demonstrate that we can use the classifier guidance technique~\cite{dhariwal2021diffusion} to steer the backward diffusion process towards classes not present in the training set of a diffusion model. Then, additionally motivated by the finding that examples located close to the decision boundary are more likely to be forgotten~\citep{toneva2018empirical},
we guide the diffusion model trained on data from preceding tasks towards recently encountered classes in generative rehearsal scenario.
The visualization of this idea is presented in Fig.~\ref{fig:teaser}.

In our experiments, we show that rehearsal with \ours{}
outperforms other state-of-the-art generative replay methods, significantly reducing catastrophic forgetting in class-incremental learning.
On top of our method, we thoroughly evaluate several alternative guidance strategies that generate rehearsal samples of diverse characteristics.
Our contributions can be summarized as follows:
\begin{itemize}
    \item We introduce \ours{} - generative replay method that benefits from classifier guidance to generate rehearsal data samples prone to be forgotten.
    \item We demonstrate that incorporating classifier guidance enables the generation of high-quality samples situated near task decision boundaries. This approach effectively mitigates forgetting in class-incremental learning.
    \item We show the superiority of \ours{} over recent state-of-the-art generative rehearsal approaches and provide an in-depth experimental analysis of our method's main contribution.
\end{itemize}

\section{Related work}
\subsection{Continual Learning}
Continual learning methods aim to mitigate catastrophic forgetting -- a phenomenon where deep neural networks trained on a sequence of tasks completely and abruptly forget previously learned information upon retraining on a new task. Recent methods can be organized into three main families. \textbf{Regularization} methods~\cite{kirkpatrick2017overcoming,zenke2017continual,li2017learning} identify the most important parameters and try to slow down their changes through regularization. \textbf{Architectural} approaches~\cite{2016rusu,2017yoon,mallya2017packnet,mallya2018piggyback} change the structure of the model for each task. \textbf{Rehearsal} methods replay data samples from previous tasks and train the model on a combination of data samples from previous and current tasks. In the most straightforward rehearsal approach, a memory buffer is used to store exemplars from previous tasks~\cite{prabhu2020gdumb,rebuffi2017icarl,chaudhry2018riemannian,wu2019large,hou2019learning,belouadah2019il2m,castro2018end}.
Some methods, instead of directly using exemplars, stores data representations from previous tasks in different forms, e.g., \emph{mnemonics} - optimized artificial samples~\cite{liu2020mnemonics}, distilled datasets~\cite{wang2018dd1,zhao2021dd2,zhao2021dd3}, or addressable memory structure~\cite{deng2022remember}.

\paragraph{Continual learning with generative rehearsal}
Because of the limitations of buffer-based approaches related to the constantly growing memory requirements and privacy issues, 
\citet{shin2017continual} introduce Deep Generative Replay, where a GAN is used to generate rehearsal samples from previous tasks for the continual training of a classifier.
A similar approach is further extended to different model architectures like Variational Autoencoders (VAE) \cite{van2018generative, nguyen2017variational}, normalizing flows \cite{scardapane2020pseudo} or Gaussian Mixture Models \cite{rostami2019complementary}.

On top of those baseline approaches, \citet{ramapuram2020lifelongvae} introduce a method that benefits from the knowledge distillation technique in VAE training in CL setup, while \citet{wu2018memory} introduce Memory Replay GANs (MeRGANs) and describe two approaches to prevent forgetting - by joint retraining and by aligning the replay samples. 

Instead of replaying images, several approaches propose to rehearse internal data representations instead, e.g., 
Brain-Inspired Replay (BIR) \cite{van2020brain} with an extension to Generative Feature Replay (GFR) \cite{liu2020generative}, where the rehearsal is combined with features distillation. \citet{kemker2017fearnet} divide feature rehearsal into short and long-term parts.

\subsection{Guided image generation in diffusion models}
Besides conditioning, controlling diffusion model outputs can be achieved by modifying the process of sampling that incorporates additional signals from the guidance function. Classifier guidance introduced by \citet{dhariwal2021diffusion} employ a trained classifier model to steer the generation process via gradients of a specified loss function, typically assessing the alignment of generated images with specific classes.
This concept is extended by \citet{bansal2023universal} to include any off-the-shelf model in guidance and further applied by \citet{augustin2022diffusion} for explaining the decisions of the classifier with generated counterfactual examples. Contrasting with these methods,  
\citet{epstein2023diffusion} introduce self-guidance based on the internal representations of the diffusion model, while \citet{ho2022classifierfree} introduce classifier-free guidance, achieving results akin to classifier-based approaches by joint training of unconditional and conditional diffusion models.

\subsection{Continual learning with diffusion models}
Diffusion models excel in generative tasks, surpassing VAEs~\cite{kingma2014autoencoding} and GANs, yet their adoption in CL remains limited. 
Deep Diffusion-based Generative Replay (DDGR) \cite{gao2023ddgr} uses a diffusion model in a generative rehearsal method and benefits from a classifier pretrained on previous tasks to synthesize high-quality replay samples. Class-Prototype Conditional Diffusion Model (CDPM) \cite{doan2023class} further enhances the replay samples quality by conditioning the diffusion model on learnable class-prototypes.
In this work, we further extend those ideas and show that classifier guidance might be used not only to generate high-quality samples but also to introduce the desired characteristics of rehearsal samples.

\section{Background}
\subsection{Diffusion models}
Let $q(\mathbf{x}_0)$ be the real data distribution. The forward diffusion process $q$ gradually adds Gaussian noise to each data sample $\mathbf{x}_0 \sim q(\mathbf{x}_0)$ over a sequence of $T$ steps according to the variance schedule $\{\beta_{t}\in(0,1)\}_{t=1}^{T}$. If we additionally define \( \alpha_t = 1 - \beta_t\) and \( \bar{\alpha}_t = \prod_{i=1}^{t}\alpha_i \), then:
\begin{equation}
    q(\mathbf{x}_t|\mathbf{x}_{t-1}) = \mathcal{N}(\mathbf{x}_t; \sqrt{\alpha_t} \mathbf{x}_{t-1}, (1-\alpha_t) \mathbf{I}).
\end{equation}
This process produces a sequence of noisy samples $\mathbf{x}_{1},\dots,\mathbf{x}_{T}$, where $\mathbf{x}_T$ is pure Gaussian noise.
The reverse diffusion process $p$ can be modeled as a diagonal Gaussian:
\begin{equation}
    p_\theta(\mathbf{x}_{t-1} | \mathbf{x}_t) = \mathcal{N}(\mathbf{x}_{t-1}; \mu_\theta(\mathbf{x}_t, t), \Sigma_\theta(\mathbf{x}_t, t)),
\end{equation}
where $\mu_\theta(\mathbf{x}_t, t)$ can be parameterized by a deep neural network $\epsilon_{\theta} (\mathbf{x}_t, t)$ with parameters $\theta$ that predicts the added noise $\epsilon_t$ for each noise level $t$:
\begin{equation}
        \mu_\theta(\mathbf{x}_t, t) = \frac{1}{\sqrt{\alpha_t}} \left(
    \mathbf{x}_t - \frac{1-\alpha_t}{\sqrt{1-\bar{\alpha}_t}}\epsilon_\theta(\mathbf{x}_t, t) \right).
\end{equation}

Sampling from DDPM requires denoising the sample sequentially from $\mathbf{x}_T \sim \mathcal{N}(0, \mathbf{I})$ $T$ steps up until $\mathbf{x}_0$. In this work, we benefit from the speed-up technique known as Denoising Diffusion Implicit Models (DDIM) introduced by \citet{song2020denoising}, where at each time step $t$, we first predict the denoised sample at $t=0$:
\begin{equation}
    \label{eq:ddim_clean_img}
    \hat{\mathbf{z}}_0(\mathbf{x}_t) = \frac{\mathbf{x}_{t}-\sqrt{1-\bar{\alpha}_{t}}\epsilon_\theta (\mathbf{x}_t, t)}{\sqrt{\bar{\alpha}_{t}}},
\end{equation}
then we calculate \(\mathbf{x}_{t-1}\) based on this prediction. This allows us to sample using fewer steps in the reverse process, reducing the inference time significantly.

\subsection{Classifier guidance}
Classifier guidance \cite{dhariwal2021diffusion} enables steering the backward diffusion process by combining the intermediate denoising steps of conditional or unconditional diffusion with the gradient from the externally trained classifier. To ensure the high quality of generated samples, the original method was introduced with a classifier trained on noised images. However, such an approach can be impractical and lead to limited performance. 

Building upon the work by \citet{bansal2023universal}, in our method, we adapt the guidance process to utilize a classifier trained only on clean images. To that end, in the guidance process, we first predict denoised image $\hat{\mathbf{z}}_0(\mathbf{x}_t)$, as in Eq.~\ref{eq:ddim_clean_img}. We can then modify the prediction of diffusion model $\epsilon_\theta (\mathbf{x}_t, t)$ at each time step $t$ according to:
\begin{equation}
    \label{eq:guidance}
    {\hat{\epsilon}}_{\theta}(\mathbf{x}_t,t)\,=\,\epsilon_{\theta}(\mathbf{x}_t,t)\,+\,s\nabla_{\mathbf{x}_{t}}{\ell}\left(f_\phi(y|\hat{\mathbf{z}}_0(\mathbf{x}_t)), y\right),
\end{equation}
where $s$ is gradient scale, $f_\phi (y|\mathbf{x})$ is classifier model with parameters $\phi$, $\ell$ is the cross-entropy loss function and $y$ is class label that we guide to.

\subsection{Continual learning with generative replay}
In this work, we focus on class-incremental continual learning \cite{van2022three} of a classifier $f_\phi(y|\mathbf{x})$ with generative replay, where we use diffusion model $\epsilon_\theta(\mathbf{x}_t,t,y)$ as a generator of synthetic samples from previous tasks. The data stream comprises of $T$ distinct tasks. Each task $i$ has its associated dataset $\mathcal{D}_i$ with collection of $N$ pairs $\{(\mathbf{x}_i^j, y_i^j) \}_{j=1}^N$, where $\mathbf{x}_i^j$ and $y_i^j$ represent the $j$-th input sample and its corresponding label, respectively. 

In each task $i \in [1, \dots, T]$
we train a classifier $\currclassifier$ that we optimize towards best performance on all preceding and current tasks,
which can be achieved by minimizing the following loss function:
\begin{equation}
        \mathcal{L}_i = \sum_{k=1}^{i} \sum_{j=1}^{N} \ell ( f_{\phi_i}(\mathbf{x}_k^j), y_k^j),
\end{equation}
where $\ell$ denotes the cross-entropy loss function. However, we do not have direct access to training data from previous tasks in the CL scenario. 
Therefore, we train the classifier model on the combination of real data from the current task and generations from previous tasks generated from a diffusion model. 
To that end, the loss function in $i$-th task can be formulated as:
\begin{equation}
        \mathcal{L}_i = \sum_{k=1}^{i-1} \sum_{j=1}^{N} \ell ( f_{\phi_i}(\hat{\mathbf{x}}_k^j), y_k^j) + \sum_{j=1}^{N} \ell ( f_{\phi_i}(\mathbf{x}_i^j), y_i^j),
\end{equation}
where as $\hat{\mathbf{x}}_k^j$ we denote the $j$-th generation from $k$-th task. Each generation is sampled from the diffusion model trained on the previous task $i-1$.

In this work, we evaluate how to generate rehearsal examples from the diffusion model to maximize the overall performance of a continually trained classifier. We do not apply any modifications to the generative replay in DDPM itself.

\section{Method}
This section introduces \ours{} - a novel method designed to mitigate catastrophic forgetting in a classifier trained in a generative replay scenario with a diffusion model. We benefit from the classifier trained on the current task, referred to as the \emph{current classifier}, to guide the diffusion model during the generation of rehearsal examples from preceding tasks. With our method, we can generate rehearsal examples close to the classifier's decision boundary, making them highly valuable to counteract the classifier's forgetting in class-incremental learning \citep{toneva2018empirical, kumari2022retrospective}.

\subsection{Intuition and rationale behind \ours{}}
\label{sec:intuition}
Before moving to the continual-learning setup, we demonstrate the effect of guiding the diffusion model towards classes not included in its training dataset. To that end, we propose a simplified scenario in which we employ the unconditional diffusion model $\epsilon_{\theta}(\mathbf{x}_t,t)$ trained exclusively on the \textit{goldfish} and \textit{tiger shark} classes from the ImageNet100 dataset, along with a classifier $f_{\phi}(y|\mathbf{x})$ trained on entire ImageNet dataset. 

\begin{figure}[ht]
\vskip -0.1in
\centering
\includegraphics[width=\columnwidth]{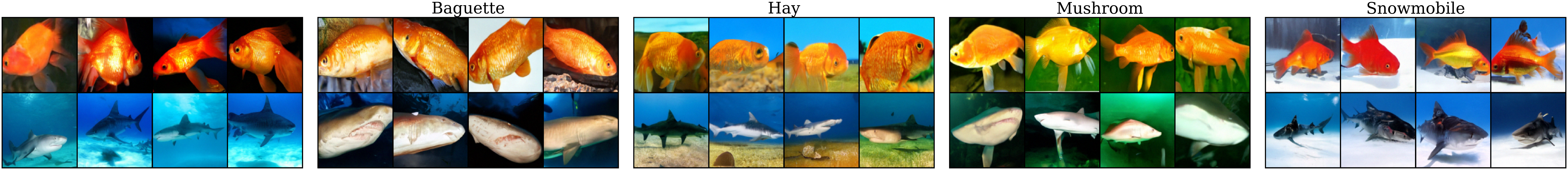}
\caption{Samples from the unconditional diffusion model trained only on \textit{goldfish} and \textit{tiger shark} classes from the ImageNet100 dataset. In the upper row, we present the samples guided to the goldfish class, while in the bottom row, to the tiger shark class. At the same time, the classifier guides the denoising process toward the class depicted above each figure that was not included in the training set of the diffusion model. For reference, in the leftmost column, we present samples generated without guidance toward any unknown class, setting $s_2 = 0$. In every other column, we set both $s_1$ and $s_2$ to $10$. We obtain samples from the desired class with observable features of classes unknown to the diffusion model, such as the color, background, or shape.}
\label{fig:toy_example}
\vskip -0.1in
\end{figure}

Given the diffusion model's unconditional nature, we use the classifier to steer the denoising process toward either the goldfish or tiger shark class, represented as $c_1$. Simultaneously, we add another guiding signal from the same classifier $f_{\phi}(y|\mathbf{x})$ towards one of the classes from the ImageNet dataset different than goldfish and tiger shark, denoted as $c_2$. Formally, we modify the unconditional diffusion sampling process as follows:
\begin{equation}
    {\hat{\epsilon}}_{\theta}(\mathbf{x}_t,t) =\epsilon_{\theta}(\mathbf{x}_t,t)
    + s_1 \nabla_{\mathbf{x}_{t}}{\ell}\left(f_{\phi} (y|\hat{\mathbf{z}}_0(\mathbf{x}_t)), c_1\right) 
    + s_2 \nabla_{\mathbf{x}_{t}}{\ell}\left(f_{\phi}(y|\hat{\mathbf{z}}_0(\mathbf{x}_t)), c_2\right).
\end{equation}
In Fig.~\ref{fig:toy_example}, we present generated samples, with goldfish generations in the upper row and tiger sharks in the bottom one.
Additionally, we present more samples generated using this technique in Appendix~\ref{app:imagenet_samples}.

By integrating guidance to both $c_1$ and $c_2$ classes, we generate samples from the diffusion model's training data distribution -- all of the examples are either goldfishes or sharks but with visible features from classes $c_2$, which are unknown to the diffusion model (e.g., oblong shape of baguette, hay or snow in the background).

In this work, we propose to use this observation in the continual training of a classifier, with the distinction that in \ours{}, we only utilize guidance toward unknown classes since we sample from a class-conditional diffusion model. Thus, we eliminate the need the need for guidance toward classes from the diffusion model's training set.

\subsection{Guidance toward classes from current task}
Our method builds upon the standard generative replay approach with a diffusion model. In each task $i$, we begin by training a classifier $\currclassifier$ on a combination of real data samples $\mathcal{D}_i$ and samples generated from the frozen previous diffusion model $\prevdiffusion$. Subsequently, we train a class-conditional diffusion model $\currdiffusion$ on currently available data $\mathcal{D}_i$ along with synthetic data samples from preceding tasks generated by the previous diffusion model $\prevdiffusion$.

When generating each rehearsal sample for classifier training, we guide the diffusion sampling process to one of the classes from the current task $i$. 

Assuming that we want to generate a replayed sample from class $y_{i-1}$, we modify the diffusion denoising process at each time step $t$ according to the Eq.~\ref{eq:guidance}, as follows:
\begin{equation}
        {\hat{\epsilon}}_{\theta_{i-1}}(\mathbf{x}_t,t,y_{i-1}) = \epsilon_{\theta_{i-1}}(\mathbf{x}_t,t,y_{i-1}) +
    s\nabla_{\mathbf{x}_{t}}{\ell}\left(f_{\phi_{i}}(y|\hat{\mathbf{z}}_0(\mathbf{x}_t), y_{i}\right).
\end{equation}

Since task $i$ can contain many classes, in each denoising step $t$, we select class $y_i$ from a current task that at that moment yields the highest output from the classifier:
\begin{equation}
    y_i = \underset{c \in \mathcal{C}_i}{\mathrm{argmax}}  f_{\phi_i}(y = c | \hat{\mathbf{z}}_0(\mathbf{x}_t)),
\end{equation}
where $\mathcal{C}_i$ denotes the set of classes in current task $i$.

Intuitively, this modification steers the diffusion process towards examples from the current task, as depicted in Fig.~\ref{fig:teaser}.
Simultaneously, since we utilize only the previous frozen diffusion model that is not trained on classes from the current task, we consistently obtain samples from desired class $y_{i-1}$. 
Rehearsal examples obtained with the modified sampling process yield lower outputs for class $y_{i-1}$ in the current classifier $\currclassifier$ compared to the previous classifier $\prevclassifier$. Hence, these examples can be interpreted as data samples that are more likely to be forgotten during continual training.  We experimentally validate this statement in Sec.~\ref{sec:guidance_analysis}.
The effect of our guidance technique resembles the idea introduced in Retrospective Adversarial Replay (RAR) \citep{kumari2022retrospective}, where authors show that training the classifier on rehearsal samples similar to examples from the current task helps the model to learn the boundaries between tasks. This is also in line with observation by~\citet{toneva2018empirical}, who show that the sample's distance from the decision border is correlated with the number of forgetting events.

\section{Experiments}
\subsection{Experimental setup}
\label{sec:exp_setup}

\paragraph{Datasets}
We evaluate our approach on CIFAR-10 and CIFAR-100~\citep{krizhevsky2009learning} image datasets.
We split the CIFAR-10 dataset into 2 and 5 equal tasks and the CIFAR-100 dataset into 5 and 10 equal tasks. Moreover, to validate if our method can be extended to datasets with higher resolution, we also evaluate it on the ImageNet-100~\cite{deng2009imagenet} dataset split into 5 tasks.

\paragraph{Metrics} For evaluation on each task $i \in [1,\dots,T]$, we use two metrics commonly used in continual learning: average accuracy \( \bar{A}_i = \frac{1}{i} \sum_{j=1}^{i} A_j^i \) and average forgetting \( \bar{F}_i = \frac{1}{i-1} \sum_{j=1}^{i-1} \max_{1 \leq k \leq i}  (A_j^k - A_j^i)\), where $A_j^i$ denotes the accuracy of a model on $j$-th task after training on $i$ tasks. We follow the definitions of both metrics from~\cite{chaudhry2018riemannian}.

\begin{table*}[t]
\caption{Comparison of \ours{} with other generative rehearsal methods (we mark feature replay methods in gray color). Our approach outperforms most other methods in terms of both average accuracy and average forgetting after the final task $T$.}
\label{tab:main_results}
\centering
\begin{sc}
\resizebox{\textwidth}{!}{ 
\begin{tabular}{lcccccccccc}
\toprule
 & \multicolumn{5}{c}{Average accuracy $\bar{A}_T$ ($\uparrow$)} & \multicolumn{5}{c}{Average forgetting $\bar{F}_T$ ($\downarrow$)} \\
 \toprule
\multirow{2}{*}{\textbf{Method}} & \multicolumn{2}{c}{CIFAR-10} & \multicolumn{2}{c}{CIFAR-100} & ImageNet100& \multicolumn{2}{c}{CIFAR-10} & \multicolumn{2}{c}{CIFAR-100} & ImageNet100\\
 & $T= 2$ & $T=5$ & $T=5$ & $T=10$ & $T=5$ & $T= 2$ & $T=5$ & $T=5$ & $T=10$ & $T=5$ \\
\midrule
Joint & \multicolumn{2}{c}{93.14 $\pm$ 0.16} & \multicolumn{2}{c}{72.32 $\pm$ 0.24} & 66.85  $\pm$ 2.25 & - & - & -  &  - & -\\
Continual Joint & 85.63 $\pm$ 0.39 & 86.41 $\pm$ 0.32 & 73.07 $\pm$ 0.01  & 64.15 $\pm$ 0.98 &  50.59 $\pm$ 0.35 & \phantom07.91 $\pm$ 0.67 & \phantom02.90 $\pm$ 0.08 & \phantom07.80 $\pm$ 0.55 & \phantom06.67 $\pm$ 0.36 & 12.28 $\pm$ 0.07\\
Fine-tuning & 47.22 $\pm$ 0.06 & 18.95 $\pm$ 0.20 & 16.92 $\pm$ 0.03 & 9.12 $\pm$ 0.04 & 13.49 $\pm$ 0.18 & 92.69 $\pm$ 0.06 & 94.65 $\pm$ 0.17 & 80.75 $\pm$ 0.22 & 87.67 $\pm$ 0.07 & 64.93 $\pm$ 0.00 \\
\midrule
DGR VAE & 60.24 $\pm$ 1.53 & 28.23 $\pm$ 3.84 & 19.66 $\pm$ 0.27 &10.04 $\pm$ 0.17 & 9.54 $\pm$ 0.26 & 43.91 $\pm$ 5.40 & 57.21 $\pm$ 9.82& 42.10 $\pm$ 1.40 & 60.31 $\pm$ 4.80 & 40.26 $\pm$ 0.91 \\
DGR+distill & 52.40 $\pm$ 2.58 & 27.83 $\pm$ 1.20 & 21.38 $\pm$ 0.61  & 13.94 $\pm$ 0.13 &11.77 $\pm$ 0.47 & 70.84 $\pm$ 6.35 & 43.43 $\pm$ 2.60 & 29.30 $\pm$ 0.40 &21.15 $\pm$ 1.30 & 41.17 $\pm$ 0.43\\
RTF & 51.80 $\pm$ 2.56 & 30.36 $\pm$ 1.40 & 17.45 $\pm$ 0.28 & 12.80 $\pm$ 0.78 & 8.03 $\pm$ 0.05 & 60.49 $\pm$ 5.54 & 51.77 $\pm$ 1.00& 47.68 $\pm$ 0.80 & 45.21 $\pm$ 5.80 & 41.2 $\pm$ 0.20\\
MeRGAN & 50.54 $\pm$ 0.08  & 51.65 $\pm$ 0.40 & \phantom09.65 $\pm$ 0.14  & 12.34 $\pm$ 0.15  & - & - & - & - & - & -\\
\rowcolor{lightergray} BIR & 53.97 $\pm$ 0.97 & 36.41 $\pm$ 0.82 & 21.75 $\pm$ 0.08 & 15.26 $\pm$ 0.49 & 8.63 $\pm$ 0.19 & 64.97 $\pm$ 2.15 & 65.28 $\pm$ 1.27 & 48.38 $\pm$ 0.44 & 53.08 $\pm$ 0.75 & 40.99 $\pm$ 0.36\\
\rowcolor{lightergray} GFR & 64.13 $\pm$ 0.88 & 26.70 $\pm$ 1.90 & 34.80 $\pm$ 0.26 & 21.90 $\pm$ 0.14 & 32.95 $\pm$ 0.35 & 25.37 $\pm$ 6.62 & 49.29 $\pm$ 6.03 & \textbf{19.16} $\pm$ 0.55 & \textbf{17.44} $\pm$ 2.20 & \textbf{20.37} $\pm$ 1.47 \\
DDGR & 80.03 $\pm$ 0.65 & 43.69 $\pm$ 2.60 & 28.11 $\pm$ 2.58 & 15.99 $\pm$ 1.08 & 25.59 $\pm$ 2.29 & 22.45 $\pm$ 1.13 & 62.51 $\pm$ 3.84 & 60.62 $\pm$ 2.13 & 74.70 $\pm$ 1.79 & 49.52 $\pm$ 2.52 \\
DGR diffusion & 77.43 $\pm$ 0.60 & 59.00 $\pm$ 0.57 & 28.25 $\pm$ 0.22 & 15.90 $\pm$ 1.01 & 23.92 $\pm$ 0.92 & 26.32 $\pm$ 0.90 & 40.38 $\pm$ 0.32 & 68.70 $\pm$ 0.65 & 80.38 $\pm$ 1.34 & 54.44 $\pm$ 0.14\\

\textbf{\ours} & \textbf{81.29} $\pm$ 0.75 & \textbf{64.47} $\pm$ 0.45 & \textbf{41.66} $\pm$ 0.40 & \textbf{26.13} $\pm$ 0.29 & \textbf{39.07} $\pm$ 1.37 & \textbf{14.79} $\pm$ 0.36 & \textbf{24.84} $\pm$ 0.05 & 44.30 $\pm$ 1.10 & 60.54 $\pm$ 0.82 & 27.60 $\pm$ 3.28 \\

\bottomrule
\end{tabular}
} 
\end{sc}
\vskip -0.2in
\end{table*}

\paragraph{Baseline methods}
We compare our approach with state-of-the-art generative replay methods. 
For fairness, in the evaluation of \textbf{BIR} \citep{van2020brain}, we freeze the encoder model after training on the first task. In order to evaluate \textbf{MeRGAN} \cite{wu2018memory} on generative replay scenario, we generate samples from the final generator network trained sequentially on all tasks to construct a training dataset for the ResNet18 classifier.
We also compare the results to the \textbf{Joint} training on all data and simple \textbf{Fine-tuning} with no rehearsal. 
As a soft upper bound of the proposed method, we present a \textbf{Continual Joint} setting, where we train the classifier continually with full access to all previous tasks (perfect rehearsal with infinite buffer size).
In all methods using the diffusion model (including \textbf{DDGR}), we use the same number of denoising steps to obtain rehearsal samples. We recalculated scores for all related methods using the code provided by the authors. Importantly, we do not use any pre-training on external datasets.

\subsection{Implementation details}
\label{sec:implementation_details}
Our training procedure is divided into two parts. First, in each task $i$, we train the classifier $\currclassifier$. In order to generate rehearsal samples, we load the diffusion model $\prevdiffusion$ already trained on the previous task. Then, we train a class-conditional diffusion model $\currdiffusion$ in a standard self-rehearsal approach, independently of the classifier. In the first task, we train both the diffusion model and the classifier solely on real data samples.

In our method, it is essential to generate rehearsal examples progressively as the classifier is trained on the task. This way, similarly to active learning techniques, at each step, we guide the generation process toward the most challenging samples -- those that are close to the decision boundary of the classifier at the given moment. The mini-batches are balanced to ensure that each contains an equal number of samples from each class encountered so far. 
For clarity, we present a detailed pseudocode of the entire training procedure in Appendix~\ref{app:pseudocode}.
All training hyperparameters can be found in Appendix~\ref{app:training_details} and in the code repository\footnote{\href{https://github.com/cywinski/guide}{\texttt{https://github.com/cywinski/guide}}}.

\subsection{Experimental results}
\label{sec:exp_results}

\begin{wrapfigure}{R}{0.55\textwidth}
\vskip -0.3in
\centering
\includegraphics[width=0.55\textwidth]{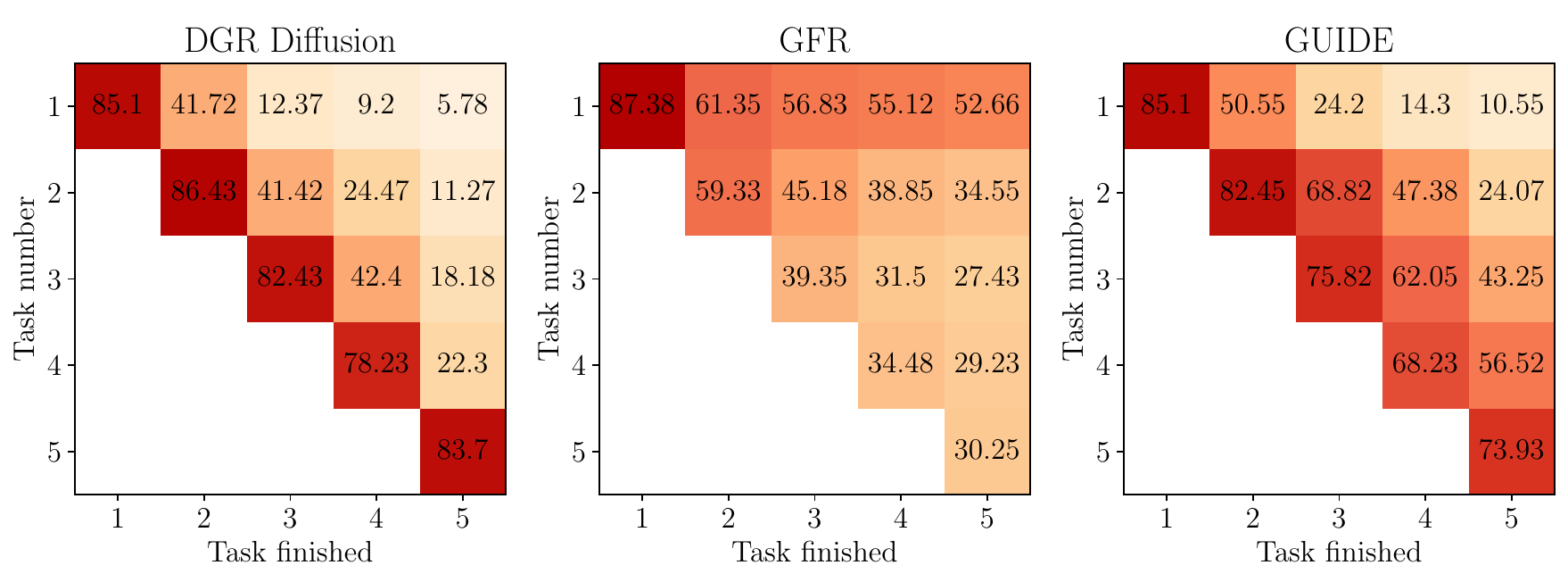}
\caption{Accuracy on each task during each phase of class-incremental training on CIFAR-100 with 5 tasks - standard GR with diffusion (left), GFR (middle), and our method (right). We observe the stability-plasticity trade-off, where our method significantly reduces forgetting compared to the standard GR scenario at the cost of a slight decrease in the ability to learn new tasks.}
\label{fig:heatmap}
\end{wrapfigure}

We evaluate our method on a set of CL benchmarks with a comparative analysis conducted in relation to other generative rehearsal techniques. In Tab.~\ref{tab:main_results}, we present the mean and standard deviation of results calculated for 3 random seeds. Our method outperforms other evaluated methods regarding the average accuracy after the last task $\bar{A}_T$ by a considerable margin on all benchmarks. Specifically, \ours{} notably improves upon the standard DGR with diffusion model on both average incremental accuracy and forgetting. It also outperforms DDGR, another GR approach that uses the diffusion model, proving the superiority of our sampling technique.

We further compare our method to feature replay methods (BIR and GFR). In the case of the GFR, we observe less forgetting on the CIFAR-100 and ImageNet100 datasets. This is the effect of the drastically limited plasticity of GFR due to the decrease in the number of updates to the feature extractor after the first task. For a more detailed analysis, in Fig.~\ref{fig:heatmap}
we thoroughly benchmark our method against the standard GR scenario and GFR method, presenting the accuracy on each encountered task after each training phase on CIFAR-100 with 5 tasks. Our approach significantly improved upon the standard GR scenario in terms of knowledge retention from preceding tasks. 
It indicates that training on rehearsal examples generated by \ours{} successfully mitigates forgetting.
The decreased accuracy on the most recent task in our method can be interpreted through the lens of the stability-plasticity trade-off \citep{grossberg1982studies}, highlighting that while our approach substantially reduces the forgetting of classifier, it does so at the expense of its ability to assimilate new information. Although GFR can maintain the performance on the initial task notably better than our method thanks to limited training of feature extractor on subsequent tasks, it leads to a substantial final performance drop.

\subsection{Analysis of the proposed guidance}
\label{sec:guidance_analysis}

\begin{wraptable}{R}{0.4\textwidth}
\vskip -0.4in
\caption{Proportion of misclassified rehearsal samples after the perturbation. Samples generated via \ours{} exhibit a higher misclassification rate, signifying their proximity to the classifier's decision boundary. Moreover, rehearsal samples in our method yield lower outputs for both previous and current classifiers.}
\label{tab:missclassified}
\centering
\begin{sc}
\resizebox{0.4\textwidth}{!}{ 

\begin{tabular}{lccc}
\toprule
 &misclassified & \multicolumn{2}{c}{confidence} \\
 &examples & prev & curr \\ 
\midrule
DGR diffusion & 55.13\% & 99.6\% & 90.03\% \\
\ours{} & 72.66\% & 86.42\% & 61.61\% \\
\bottomrule
\end{tabular}
} 

\end{sc}
\end{wraptable}

To demonstrate that our method produces samples near the decision boundary of a classifier, we propose to launch a simple adversarial attack on the generated samples in order to check how easy it is to change their class to the one from the current task.
Concretely, we adapt the method introduced by \citet{DBLP:journals/corr/GoodfellowSS14}, and modify each rehearsal sample that was generated during training $\hat{\mathbf{x}}$ as follows:
\begin{equation}
       \hat{\mathbf{x}}^{*} = \hat{\mathbf{x}} - \epsilon  \text{sign}\left( \nabla_{\hat{\mathbf{x}}} \ell(f_{\phi_{i}}(y|\hat{\mathbf{x}}), y_i) \right), 
\end{equation}
where $\epsilon=0.1$ and $y_i = \underset{c \in \mathcal{C}_i}{\mathrm{argmax}} \; f_{\phi_i}(y = c | \hat{\mathbf{x}})$.

Then, we calculate the proportion of cases where the classifier's prediction for the modified sample $\hat{\mathbf{x}}^{*}$ differs from its prediction for the original generated sample $\hat{\mathbf{x}}$. As shown in Tab.~\ref{tab:missclassified}, we can change the classifier's prediction much more frequently when we sample the replay examples according to our method. Since rehearsal examples generated with \ours{} are much more likely to be misclassified after a simple modification with predefined magnitude, this indicates that the modification of diffusion's prediction introduced in our method successfully moves the generations closer to the classifier's decision boundary.

\begin{wrapfigure}{R}{0.5\textwidth}
\vskip -1.2in
\centering
\includegraphics[width=0.5\textwidth]{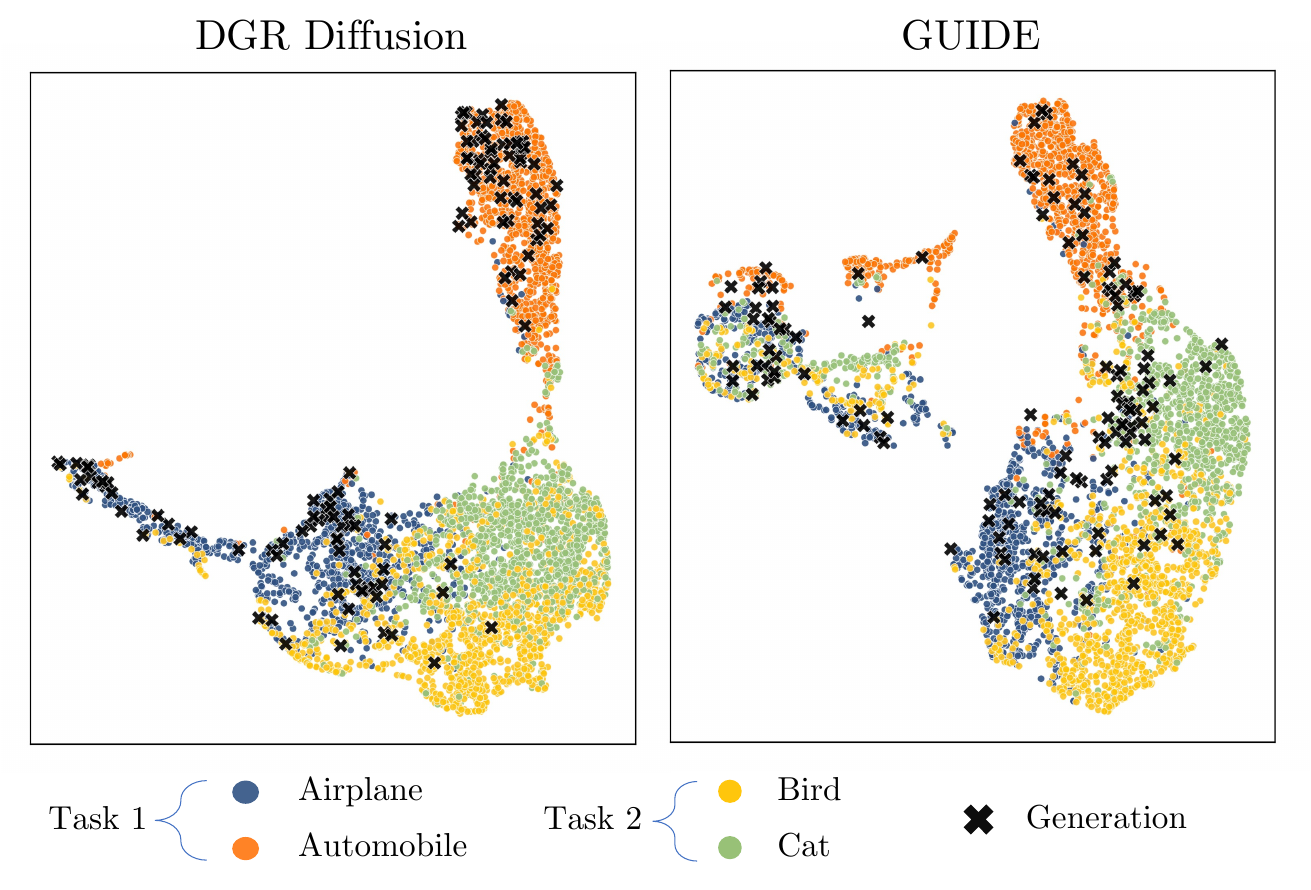}
\caption{Visualization of the classifiers embedding space (umap) for training examples and generations sampled with standard generative replay method (left) and ours (right) at 75\% of the training on the second task. We can observe how \ours{} sample generations are more similar to the training examples from new classes (e.g., airplanes similar to birds).}
\label{fig:latent_generations}
\vskip -0.2in
\end{wrapfigure}

Moreover, we present a visualization of generated samples in the latent space of a classifier (Fig.~\ref{fig:latent_generations}) that we calculate during the training of the second task on a CIFAR-10 dataset divided into five equal tasks. In the standard GR scenario, the rehearsal samples originate predominantly from high-density regions of class manifolds, which is evident from their central location within each class's manifold. On the other hand, our method yields generations that are more similar to the examples from the second task.

\section{Additional analysis}
\subsection{The effect of changing classifier scale}

An important hyperparameter of our method is the gradient scaling parameter $s$ that controls the strength of the guidance signal. In this section, we show the effect of the gradient scale $s$ on the effectiveness of our method. However, our method is robust and can work well with different values of this parameter. The detailed results with other scaling parameters are presented in Appendix~\ref{app:rehearsal_scale}.  
As we increase the scale, we observe the inflection point, after which the accuracy starts to drop. This is related to the observation that with excessively large scaling parameters,  
the quality of generated samples drops significantly.

\begin{wrapfigure}{r}{0.5\columnwidth}
\vskip -0.2in
\centering
\includegraphics[width=0.5\columnwidth]{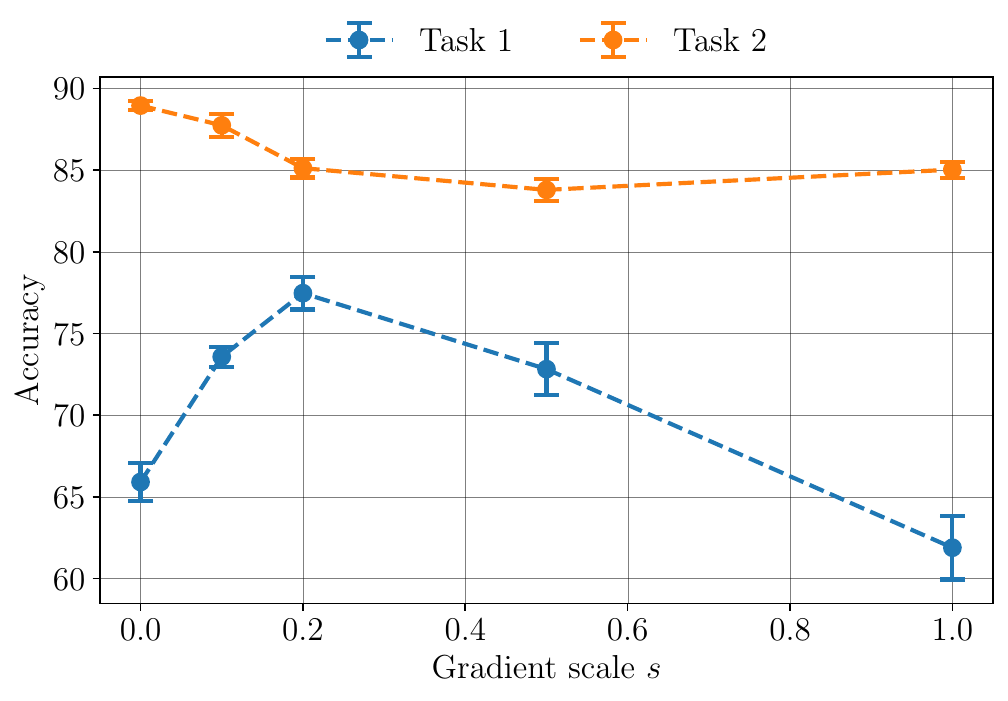}
\caption{Classifier scale impact on forgetting and ability to acquire new information. Up to $s=0.2$, when we increase scale, we reduce the forgetting but also observe a drop in the accuracy on the second task. When we use too large scale $s$, the quality of samples drops significantly, along with the accuracy on the previous task. We further present this effect in Appendix~\ref{app:rehearsal_scale}.}
\label{fig:acc_prev_tasks}
\end{wrapfigure}

In Fig.~\ref{fig:acc_prev_tasks}, we present results averaged over 3 random seeds for different values of gradient scale $s$, presenting standard deviation as error bars. We observe that the scaling parameter introduces a trade-off between the stability and plasticity of the continually trained classifier. When we increase $s$, the accuracy on the previous task increases along with a drop in accuracy on the current task.

\subsection{Alternative variants of guidance}
In addition to our primary method, we evaluate alternative variants of incorporating classifier guidance to generative replay setup, drawing inspiration from corresponding techniques in buffer-based rehearsal. Each variant effectively modifies the sampling strategy from the diffusion model. In each variant, we benefit either from the frozen classifier $\prevclassifier$, trained on prior tasks and henceforth referred to as the \emph{previous classifier}, or the currently trained classifier, $\currclassifier$. In this section, we define each variant highlighting with the \textcolor{blue}{blue} color the distinctions from \ours{}. 

\begin{table*}[t!]
\caption{Comparison of evaluated variants of integrating classifier guidance in CL. \textbf{PREV} and \textbf{CURR} refers to guidance from previous and current classifier respectively. Guidance toward classes from the previous tasks is denoted with "-" and guidance towards classes from the current task with "+". Each of the introduced variants outperforms standard DGR with diffusion model on most of the evaluated benchmarks and achieves state-of-the-art performance. }
\label{tab:variants_comparison}
\centering
\begin{sc}
\resizebox{\textwidth}{!}{ 
\begin{tabular}{lcccccccccc}
\toprule
 & \multicolumn{5}{c}{Average accuracy $\bar{A}_T$ ($\uparrow$)} & \multicolumn{5}{c}{Average forgetting $\bar{F}_T$ ($\downarrow$)} \\
\toprule
\multirow{2}{*}{\textbf{Variant}} & \multicolumn{2}{c}{CIFAR-10} & \multicolumn{2}{c}{CIFAR-100} & ImageNet100 & \multicolumn{2}{c}{CIFAR-10} & \multicolumn{2}{c}{CIFAR-100} & ImageNet100\\
  & $T= 2$ & $T=5$ & $T=5$ & $T=10$ & $T=5$ & $T= 2$ & $T=5$ & $T=5$ & $T=10$ & $T=5$ \\
\midrule
DGR diffusion & 77.43 $\pm$ 0.60 & 59.00 $\pm$ 0.57 & 28.25 $\pm$ 0.22 & 15.90 $\pm$ 1.01 & 23.92 $\pm$ 0.92 & 26.32 $\pm$ 0.90 & 40.38 $\pm$ 0.32 & 68.70 $\pm$ 0.65 & 80.38 $\pm$ 1.34 & 54.44 $\pm$ 0.14\\
PREV + & 80.03 $\pm$ 0.65 & 60.31 $\pm$ 0.44 & 31.35 $\pm$ 0.66 & 18.22 $\pm$ 0.52 & 29.13 $\pm$ 2.81 & 22.45 $\pm$ 1.13 & 40.00 $\pm$ 0.60 & 64.80 $\pm$ 1.00 & 77.60 $\pm$ 0.50 & 45.15 $\pm$ 3.64 \\
PREV -& 75.79 $\pm$ 1.20 & 57.05 $\pm$ 0.43 & 28.40 $\pm$ 0.04 & 15.79 $\pm$ 0.21 & 12.60 $\pm$ 0.71 & 27.58 $\pm$ 0.95 & 44.80 $\pm$ 0.90 & 68.34 $\pm$ 0.29 & 80.42 $\pm$ 0.44 & 59.98 $\pm$ 0.28 \\
CURR - & 78.72 $\pm$ 0.58 & 57.72 $\pm$ 0.95 & 30.57 $\pm$ 0.33 & 16.87 $\pm$ 0.83 & 23.55 $\pm$ 3.27 & 23.89 $\pm$ 0.44 & 43.31 $\pm$ 1.50 & 65.42 $\pm$ 0.81 & 78.96 $\pm$ 1.16 & 53.85 $\pm$ 3.60 \\
\textbf{\ours} & \textbf{81.29} $\pm$ 0.75 & \textbf{64.47} $\pm$ 0.45 & \textbf{41.66} $\pm$ 0.40 & \textbf{26.13} $\pm$ 0.29 & \textbf{39.07} $\pm$ 1.37 & \textbf{14.79} $\pm$ 0.36 & \textbf{24.84} $\pm$ 0.05 & \textbf{44.30} $\pm$ 1.10 & \textbf{60.54} $\pm$ 0.82 & \textbf{27.60} $\pm$ 3.28 \\
\bottomrule
\end{tabular}
} 
\end{sc}
\vskip -0.2in
\end{table*}

\paragraph{Guidance towards classes from previous tasks}
The most straightforward adaptation of a classifier guidance concept to a generative replay setup is to modify the diffusion sampling process using the gradients from the previous frozen classifier to refine the quality of rehearsal samples. The modification of the previous diffusion model's prediction can be thus defined as:
\begin{equation}
        {\hat{\epsilon}}_{\theta_{i-1}}(\mathbf{x}_t,t,y_{i-1}) =  \epsilon_{\theta_{i-1}}(\mathbf{x}_t,t,y_{i-1}) +
    s\nabla_{\mathbf{x}_{t}}{\ell}\left(\textcolor{blue}{f_{\phi_{i-1}}}(y|\hat{\mathbf{z}}_0(\mathbf{x}_t)), \textcolor{blue}{y_{i-1}}\right),
\end{equation}
where $y_{i-1}$ denotes the class label from one of the previous tasks. As noted by~\cite{dhariwal2021diffusion}, the application of classifier guidance creates a trade-off: it enhances the quality of the generated samples at the cost of their diversity. This approach is similar to the one introduced by \citet{gao2023ddgr} except that we do not use guidance in the process of continual diffusion training but only in the classifier's training. Intuitively similar buffer-based methods are based on the herding algorithm~\citep{welling2009herding} and used in iCaRL method~\citep{rebuffi2017icarl}, which seeks to store samples that best represent the mean of classes in the feature space.

\paragraph{Guidance away from classes from previous tasks}
Alternatively, we can guide the diffusion-denoising process in the opposite direction by
maximizing the entropy of the classifier instead of minimizing it. As noted by \citet{sehwag2022generating}, such an approach steers the denoising diffusion process away from the high-density regions of the data manifold. Consequently, it should generate synthetic samples that resemble the rare instances in the training dataset, which are typically more challenging for the classifier to identify.
In the first variant, we propose to use the previous classifier $\prevclassifier$ to guide away from the old classes:
\begin{equation}
        {\hat{\epsilon}}_{\theta_{i-1}}(\mathbf{x}_t,t,y_{i-1}) = \epsilon_{\theta_{i-1}}(\mathbf{x}_t,t,y_{i-1}) \textcolor{blue}{-}
    s\nabla_{\mathbf{x}_{t}}{\ell}\left(\textcolor{blue}{f_{\phi_{i-1}}}(y|\hat{\mathbf{z}}_0(\mathbf{x}_t)), \textcolor{blue}{y_{i-1}}\right).
\end{equation}
Analogously, we can steer the diffusion denoising process away from the desired class from the previous task, but using the current classifier $\currclassifier$:
\begin{equation}
        {\hat{\epsilon}}_{\theta_{i-1}}(\mathbf{x}_t,t,y_{i-1}) = \epsilon_{\theta_{i-1}}(\mathbf{x}_t,t,y_{i-1})  \textcolor{blue}{-}
    s\nabla_{\mathbf{x}_{t}}{\ell}\left(f_{\phi_{i}}(y|\hat{\mathbf{z}}_0(\mathbf{x}_t)), \textcolor{blue}{y_{i-1}}\right).
\end{equation}

In both approaches, we increase the diversity
of rehearsal samples under frozen or continually trained classifiers. This variation resembles the buffer-based method of Gradient Sample Selection (GSS) \citep{zaeemzadeh2019iterative}, which seeks to maximize the diversity of the samples stored in the memory buffer.

\paragraph{Evaluation}

We evaluate the performance of each proposed variant and present the mean and standard deviation of those experiments on CL benchmarks in Tab.~\ref{tab:variants_comparison}, which are calculated for 3 random seeds. We can observe that thanks to the improved quality of rehearsal samples, approaches integrating guidance towards selected classes achieve higher overall performance. 
In simpler scenarios, increasing the diversity of generated samples can also yield slight improvement, while in more complex settings, its performance is comparable to the baseline. Nevertheless, in all evaluated scenarios, our proposed guidance towards forgotten examples outperforms competing approaches.

\section{Discussion and limitations}
\label{sec:limitations}
In this work, we focus on the continual learning of a classifier given a continually trained diffusion model. In our experiments, we train the diffusion model with simple generative replay -- a potentially suboptimal solution. As presented in Appendix~\ref{app:perfect_diff}, the significant drawback of our method is the forgetting happening in the diffusion model itself. Our initial results suggest that because of the known issue of sample deficiency from low-density regions~\cite{sehwag2022generating} forgetting in diffusion models manifests itself as a degradation of the diversity of generated samples, which has an important effect on the quality of rehearsal samples. We discuss this issue briefly in Appendix~\ref{app:diff}. Nevertheless, the forgetting of diffusion models is an open research question.
An important drawback of all generative replay approaches is the computational burden associated with training and sampling from the diffusion model. This is also true for our method, which is computationally expensive. However, \ours{} is only slightly more computationally expensive than standard DGR with diffusion and significantly less expensive than a recent state-of-the-art solution -- DDGR, while yielding better results. Moreover, we evaluate additional speed-up techniques that mitigate this burden. Detailed analysis is provided in Appendix~\ref{app:speedup}.

\section{Conclusion}
In this work, we propose \ours{}: generative replay method that utilizes classifier guidance to generate rehearsal samples that the classifier model is likely to forget. 
We benefit from a classifier trained continually in each task to guide the denoising diffusion process toward the most recently encountered classes. This strategy enables the classifier's training with examples near its decision boundary, rendering them particularly valuable for continual learning.  
Across various CL benchmarks, \ours{} demonstrates superior performance, consistently surpassing recent state-of-the-art generative rehearsal methods. This underscores the effectiveness of our approach in mitigating forgetting and training a robust classifier.

\newpage
\bibliographystyle{abbrvnat}
\bibliography{main}
\newpage
\appendix

\section{Impact of classifier scale hyperparameter on our method}
\label{app:rehearsal_scale}

To measure the effect of changing the strength of guidance scale parameter $s$ on the effectiveness of our method, we sweep over each value in $[0.1, 0.2, 0.5, 1.0]$ for each evaluated benchmark. We present the mean and standard deviation of results in Tab.~\ref{tab:acc_scales_all} calculated for 3 different values of random seed. We see that with this hyperparameter properly tuned, we are able to improve the results of our method further. Nonetheless, the results we achieved show that our method works well with different values of this hyperparameter.

\begin{table*}[ht]
\caption{Effect of gradient scale $s$ on \ours{}. By tuning this hyperparameter, we can improve the results further.}
\label{tab:acc_scales_all}
\centering
\begin{small}
\begin{sc}
\resizebox{\textwidth}{!}{ 
\begin{tabular}{ccccccccccc}
\toprule
 & \multicolumn{5}{c}{Average accuracy $\bar{A}_T$ ($\uparrow$)} & \multicolumn{5}{c}{Average forgetting $\bar{F}_T$ ($\downarrow$)} \\
\toprule
\multirow{2}{*}{\textbf{Scale $s$}} & \multicolumn{2}{c}{CIFAR-10} & \multicolumn{2}{c}{CIFAR-100} & ImageNet100-64 & \multicolumn{2}{c}{CIFAR-10} & \multicolumn{2}{c}{CIFAR-100} & ImageNet100-64\\
  & $T= 2$ & $T=5$ & $T=5$  & $T=10$& $T=5$ & $T= 2$ & $T=5$ & $T=5$ & $T=10$ &  $T=5$ \\
\midrule
0.0 & 77.43 $\pm$ 0.60 & 56.61 $\pm$ 1.85 & 28.25 $\pm$ 0.22 & 15.90 $\pm$ 1.00 & 23.92 $\pm$ 0.92  & 26.32 $\pm$ 0.90 & 43.79 $\pm$ 0.41 & 68.70 $\pm$ 0.65 & 80.38 $\pm$ 1.34 &54.44 $\pm$ 0.14 \\
0.1 & 80.66 $\pm$ 0.44 & 59.56 $\pm$ 0.52 & 31.63 $\pm$ 0.81 & 18.28 $\pm$ 0.93 & 26.48 $\pm$ 3.79 & 18.81 $\pm$ 0.48 & 39.78 $\pm$ 0.92 & 63.87 $\pm$ 1.35 & 78.01 $\pm$ 0.59 & 50.09 $\pm$ 5.04\\
0.2 & \textbf{81.29} $\pm$ 0.75 & 61.30 $\pm$ 0.10 & 37.51 $\pm$ 1.23 & 22.68 $\pm$ 0.30 & 31.09 $\pm$ 4.17 & \textbf{14.79} $\pm$ 0.36 & 34.98 $\pm$ 0.13 & 54.52 $\pm$ 1.45 & 70.68 $\pm$ 0.23 & 44.40 $\pm$ 5.02 \\
0.5 & 78.30 $\pm$ 0.47 & \textbf{64.47} $\pm$ 0.45 & \textbf{41.66} $\pm$ 0.40 & 25.48 $\pm$ 1.16 & 35.82 $\pm$ 0.56 & 20.03 $\pm$ 1.85 & 29.25 $\pm$ 1.15 & 44.30 $\pm$ 1.10 & 64.06 $\pm$ 1.53 & 35.80 $\pm$ 0.32 \\
1.0 & 73.46 $\pm$ 0.81 & 62.56 $\pm$ 0.52  & 39.55 $\pm$ 0.19 & \textbf{26.13} $\pm$ 0.29 & \textbf{39.07} $\pm$ 1.37 & 30.96 $\pm$ 1.93 & \textbf{24.05} $\pm$ 0.72 & \textbf{44.03} $\pm$ 0.34 & \textbf{60.54} $\pm$ 0.82 & \textbf{27.60} $\pm$ 3.29 \\
\bottomrule
\end{tabular}
} 
\end{sc}
\end{small}
\end{table*}

\begin{figure}[ht]
\centering
\includegraphics[width=0.8\columnwidth]{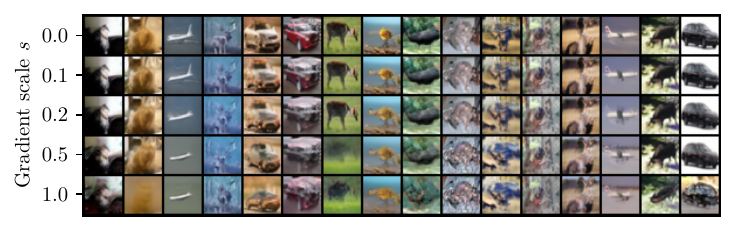}
\caption{Sample rehearsal examples of \ours{} generated by class-conditional diffusion model trained on the first task of CIFAR-10/2 setup. If we set the gradient scale parameter $s$ too large, we observe a significant drop in the quality of samples.}
\label{fig:scale_sweep}
\end{figure}

Moreover, in Fig.~\ref{fig:scale_sweep}, we present random rehearsal samples from our CIFAR-10 setup with 2 tasks generated from the same initial noise in each column. We observe the effect of classifier scale parameter $s$ on the quality of rehearsal samples in \ours{}. If we set the scale to be too large, we observe significant degradation in the quality of generations. Hence, in Tab.~\ref{tab:acc_scales_all}, we observe a drop in performance on setups where the scale is too large.

\section{Continual learning with ground-truth diffusion}
\label{app:perfect_diff}
\begin{figure}[ht!]
\centering
\includegraphics[width=.8\columnwidth]{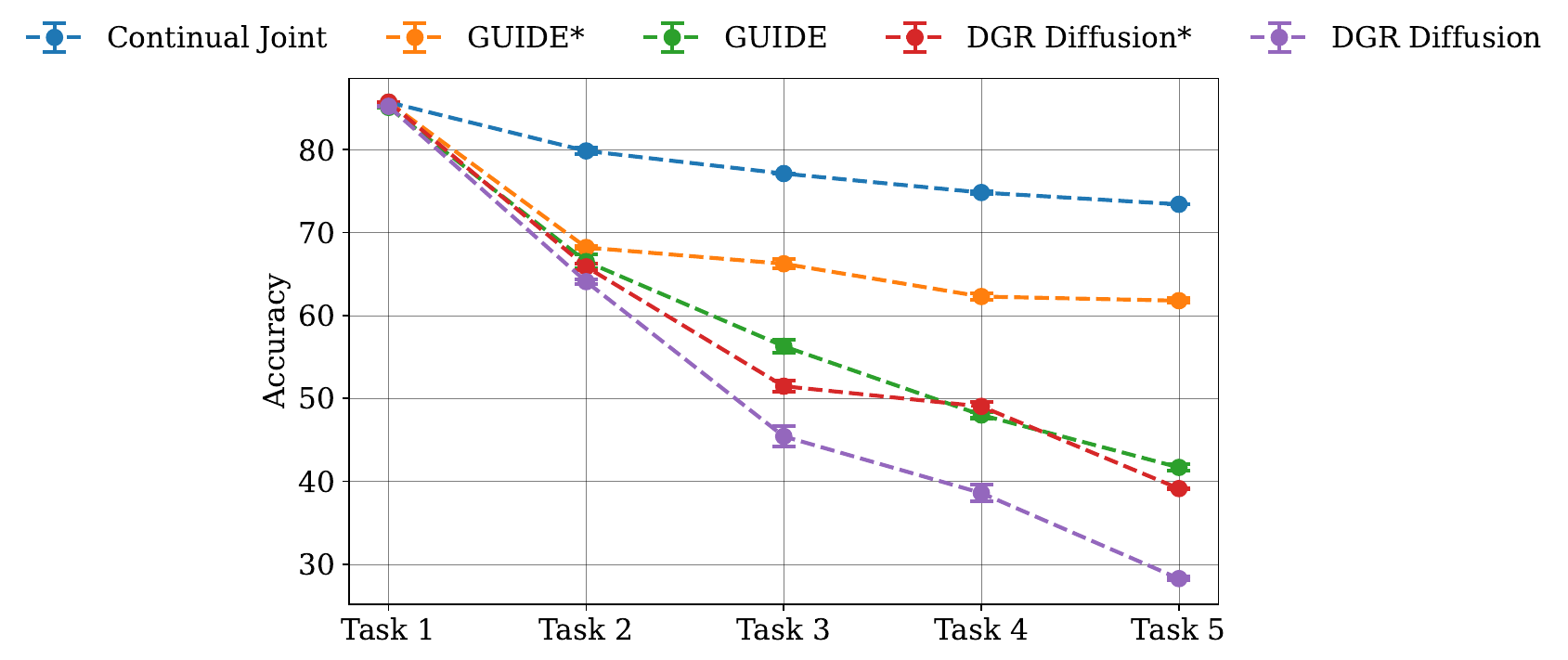}
\caption{Average accuracy on all classes seen so far after $i^{th}$ task. We highlight with asterisks $(\ast)$ methods trained with ground-truth diffusion model. We can observe that under ground-truth diffusion model \ours{} approaches the soft upper-bound defined by the continual joint training.}
\vskip -0.2in
\label{fig:perfect_diff}
\end{figure}

In all of our setups presented in the main part of the article, we trained the diffusion model alongside the continually trained classifier. In this experiment, we study the effect of the rehearsal sampling method proposed in \ours{} under the assumption of a ground-truth sampling method from the previous tasks. We independently train four diffusion models on all of the data presented up to the $i^{th}$ task in the CIFAR100/5 tasks scenario. With those models, we compare our method with the baseline sampling method and continual upper-bound where the classifier has full access to all training data samples from current and previous tasks. We depict the results of this comparison in Fig.~\ref{fig:perfect_diff}, calculated for 3 random seeds. We report mean and standard deviation as error bars. Results indicate that the random sampling of rehearsal samples, even from the ground-truth diffusion model, leads to a significant drop in performance. At the same time, our method significantly reduces the difference with the soft-upper bound defined by the continual-joint training with infinite buffer size. Interestingly, our fully continual learning setup still outperforms the baseline sampling approach rehearsed with a ground-truth diffusion. 

\newpage
\section{Pseudocode of training procedure in GUIDE}
\label{app:pseudocode}
To clarify and enhance understanding of our continual training process, this section includes pseudocode for the primary components involved in our method. Alg.~\ref{alg:guide_sample} presents a sampling of a single rehearsal example during the continual training of a classifier in \ours{}. In Alg.~\ref{alg:guide_training}, we present a complete end-to-end continual training of both diffusion model and classifier in \ours{}. 

Since the decision boundary changes during the training of a classifier, we sample new rehearsal examples according to our sampling method progressively once every $N_g$ batches. It ensures that replayed examples are located close to the decision boundary of a classifier at the given moment. Simultaneously, hyperparameter $N_g$ allows us to significantly speed up our method without a noticeable drop in accuracy, which we describe further in Appendix~\ref{app:speedup}. Since we do not use classifier guidance during the generation of a dataset for diffusion training, we sample rehearsal examples all at once and merge them with real data samples. 

In the pseudocode, consistent with the terminology used in the main text, the current classifier is denoted as $\currclassifier$ and the previous diffusion model as $\prevdiffusion$. All hyperparameters, such as the number of training steps or the number of denoising steps, are listed in Appendix~\ref{app:training_details}.

\begin{algorithm}
\caption{Rehearsal sampling in GUIDE during training on task $i$}\label{alg:guide_sample}
\begin{algorithmic}
\Require $\mathcal{C}_{i-1}$: classes from all previous tasks, $\mathcal{C}_{i}$: classes from current task, $s$: gradient scale, $T$: number of denoising steps
\State $t \gets T$
\State $y_{i-1} \sim \mathcal{U}(\mathcal{C}_{i-1})$
\State $\mathbf{x}_t \sim \mathcal{N}(0, \mathbf{I})$
\While{$t > 0$}
\State $\hat{\mathbf{z}}_0(\mathbf{x}_t) \gets \frac{\mathbf{x}_{t}-\sqrt{1-\bar{\alpha}_{t}}\epsilon_{\theta_{i-1}} (\mathbf{x}_t, t, y_{i-1})}{\sqrt{\bar{\alpha}_{t}}}$
\State $y_i \gets \underset{c \in \mathcal{C}_i}{\mathrm{argmax}}  f_{\phi_i}(y = c | \hat{\mathbf{z}}_0(\mathbf{x}_t))$
\State ${\hat{\epsilon}}_{\theta_{i-1}}(\mathbf{x}_t,t,y_{i-1}) \gets \epsilon_{\theta_{i-1}}(\mathbf{x}_t,t,y_{i-1}) +
    s\nabla_{\mathbf{x}_{t}}{\ell}\left(f_{\phi_{i}}(y|\hat{\mathbf{z}}_0(\mathbf{x}_t), y_{i}\right)$
\State $\mathbf{x}_{t-1} \gets\sqrt{\bar{\alpha}_{t-1}}\hat{\mathbf{z}}_0(\mathbf{x}_t) + \sqrt{1-\bar{\alpha}_{t-1}}{\hat{\epsilon}}_{\theta_{i-1}}(\mathbf{x}_t,t, y_{i-1})$
\State $t \gets t - 1$
\EndWhile
\Ensure $\mathbf{x}_{0}$ \Comment{Generated rehearsal sample from class $y_{i-1}$}
\end{algorithmic}
\end{algorithm}

\begin{algorithm}
\caption{Continual training in GUIDE}\label{alg:guide_training}
\begin{algorithmic}
\Require $N_c$: number of classifier training steps, $N_d$: number of diffusion training steps, $N_g$: rehearsal generation interval, $T$: number of tasks, $B$: batch size, $f_{\phi}$: classifier model, $\epsilon_\theta$: diffusion model
\For{$i \in [1, \dots, T]$}
    \State $\mathcal{D}_i \gets$ real dataset for task $i$
    \State $\epsilon_{\theta_{i}} \gets \epsilon_{\theta_{i-1}}$
    \State $f_{\phi_{i}} \gets f_{\phi_{i-1}}$
    \State $\hat{\mathcal{B}} \gets \emptyset$
    \For{$n \in [1, \dots, N_c]$} \Comment{Classifier training}
        \If{$i == 1$}
            \State $\mathcal{B} \gets B$ real samples from $\mathcal{D}_i$
        \Else
            \State $\mathcal{B} \gets \nicefrac{B}{i}$ real samples from $\mathcal{D}_i$
            \If{$n \bmod N_g == 0$}
                \State $\hat{\mathcal{B}} \gets \texttt{GUIDE\_sample}(\texttt{num\_samples} = \nicefrac{B}{i} \times (i - 1) ,f_{\phi_i}, \epsilon_{\theta_{i-1}})$
            \EndIf
        \EndIf
        \State $\mathcal{B}_c \gets \mathcal{B} \cup \hat{\mathcal{B}}$
        \State update $f_{\phi_i}$ with $\mathcal{B}_c$
    \EndFor \\
    \State $\hat{\mathcal{D}_i} \gets \emptyset$ \Comment{Construct diffusion dataset}
    \If{$i > 1$}
        \State $\hat{\mathcal{D}_i} \gets \texttt{sample\_without\_guidance}(\texttt{num\_samples}=|\mathcal{D}_{1,\dots,i-1}|, \epsilon_{\theta_{i-1}})$
    \EndIf
    \State $\mathcal{D}_d \gets \mathcal{D}_i \cup \hat{\mathcal{D}_i}$
    \State train $\epsilon_{\theta_{i}}$ on $\mathcal{D}_d$ for $N_d$ steps
\EndFor
\end{algorithmic}
\end{algorithm}

\newpage
\section{Training details and hyperparameters}
\label{app:training_details}

\subsection{Diffusion models}
In our experiments, we follow the definitions of class-conditional diffusion model architectures described in \cite{dhariwal2021diffusion}. Across all setups, we train diffusion models using AdamW optimizer with $\beta_1 = 0.9$ and $\beta_2=0.999$. The only hyperparameter that varies between tasks is the number of iterations of training. In all setups, we train models for 100K iterations on the initial task, and then on each subsequent task, we train either for 50K or 100K iterations, depending on the setup. We use DDIM with 250 steps for the generation of rehearsal samples on the ImageNet100 dataset and 1000 denoising steps for every other dataset. The only augmentation that we use during the training of diffusion models is the random horizontal flip.  

We present the most important hyperparameters for diffusion models in Tab.~\ref{tab:diffusion_hparams}.

\begin{table}[h]
    \caption{Hyperparameters for the training of diffusion models.}
    \label{tab:diffusion_hparams}

    \centering
    \resizebox{\textwidth}{!}{
    \begin{tabular}{lccccc}
    \toprule
     & CIFAR-10/2 & CIFAR-10/5 & CIFAR-100/5 & CIFAR-100/10 & ImageNet100-64/5 \\
    \midrule
    Diffusion steps & 1000 & 1000 & 1000 & 1000 & 1000  \\
    Noise schedule & linear & linear & linear & linear & linear \\
    Channels & 128 & 128 & 128 & 128 & 192 \\
    Depth & 3 & 3 & 3 & 3 & 3\\
    Channels multiple & 1, 2, 2, 2 & 1, 2, 2, 2 & 1, 2, 2, 2 & 1, 2, 2, 2 & 1, 2, 3, 4 \\
    Heads & 4&4 & 4&4 \\
    Heads channels &  &  & & & 64 \\
    Attention resolution & 16,8 & 16,8 & 16,8 & 16,8&32,16,8 \\
    BigGAN up/downsample & \xmark & \xmark & \xmark & \xmark & \cmark  \\
    Dropout & 0.1 & 0.1 & 0.1 & 0.1 & 0.1 \\
    Batch size & 256 & 256 & 256 & 256 & 100 \\
    Learning rate & 2e-4 & 2e-4 & 2e-4 & 2e-4 & 1e-4 \\
    Iterations 1\textsuperscript{st} task & 100K & 100K & 100K & 100K & 100K  \\
    Iterations other tasks & -  & 50K  & 50K & 100K& 50K \\
    Self-rehearsal denoising steps & - & 1000 & 1000 & 1000 & DDIM250 \\
    \bottomrule
    \end{tabular}
    }
\end{table}

\subsection{Classifiers}
As a classifier model, we use the same ResNet18 architecture as the GDumb method \citep{prabhu2020gdumb} in each setup, with preactivation enabled, meaning that we place the norms and activations before the convolutional or linear layers. We train models with an SGD optimizer. We list the most important hyperparameters for classifiers in Tab.~\ref{tab:clf_hparams}.

For CIFAR-10 and CIFAR-100 datasets, we define the same set of image augmentations, which include cropping, rotating, flipping, and erasing. We also apply a transformation to the brightness, contrast, saturation, and hue, followed by the normalization to the [-1, 1] range. During the training, we apply the same augmentations for both rehearsal samples and real samples from the current task.

For the ImageNet100, we do not apply any data augmentations, but we also normalize the images to the range of [-1, 1].

\begin{table}[h]
    \caption{Hyperparameters for the training of classifier models.}
    \label{tab:clf_hparams}

    \centering
    \resizebox{\textwidth}{!}{
    \begin{tabular}{lccccc}
    \toprule
     & CIFAR-10/2 & CIFAR-10/5 & CIFAR-100/5 & CIFAR-100/10 & ImageNet100-64/5 \\
    \midrule
    Batch size & 256 & 256 & 256 & 256 &  100 \\
    Learning rate 1\textsuperscript{st} task & 0.1 & 0.1 & 0.1 & 0.1  & 0.1  \\
    Learning rate other tasks & 0.01 & 0.01 & 0.05 & 0.05 & 0.001 \\
    Iterations 1\textsuperscript{st} task & 5K & 5K & 10K & 10K&  20K \\
    Iterations other tasks & 2K & 2K & 2K & 2K & 20K \\
    Rehearsal denoising steps & DDIM50 & DDIM50 & DDIM100 & DDIM100 & DDIM50 \\
    Rehearsal generation interval & 1 & 5 & 10 & 10 & 15 \\
    \bottomrule
    \end{tabular}
    }
\end{table}

\subsection{Experiments compute resources}
\label{sec:compute_resources}
In this section, we list the hardware we used for our experiments and training times on each benchmark, both for training of diffusion models in Tab.~\ref{tab:compute_resources_diff} and classifiers in Tab.~\ref{tab:compute_resources_clf}. Although we use NVIDIA A100 GPUs in our experiments due to the efficiency of training, experiments can also be reproduced on GPUs with smaller memory, as noted in Sec.~\ref{sec:runtime}.

The total compute used in a project can be estimated from the times presented in tables, considering that each experiment is calculated with three different random seeds for statistical significance. Furthermore, we need to take into account compute used in the development stage of the project.

\begin{table}[h!]
  \caption{Training times of diffusion models for each benchmark.}
  \label{tab:compute_resources_diff}
  \centering
  \begin{tabular}{lcc}
    \toprule
    Benchmark & Time[GPU-hours] & GPU used  \\
    \midrule
    CIFAR-10/2 & 5.55  & 4 x NVIDIA A100 40GB \\
    CIFAR-10/5 & 20 & 4 x NVIDIA A100 40GB \\
    CIFAR-100/5 & 20  & 4 x NVIDIA A100 40GB \\
    CIFAR-100/10 & 63.75 & 4 x NVIDIA A100 40GB \\
    ImageNet100-64/5 & 88.43 &  4 x NVIDIA A100 40GB \\
    \bottomrule
  \end{tabular}
\end{table}

\begin{table}[h!]
  \caption{Training times of classifiers for each benchmark.}
  \label{tab:compute_resources_clf}
  \centering
  \begin{tabular}{lcc}
    \toprule
    Benchmark & Time[GPU-hours] & GPU used  \\
    \midrule
    CIFAR-10/2 & 4.8 & 1 x NVIDIA A100 40GB \\
    CIFAR-10/5 & 5.8 & 1 x NVIDIA A100 40GB \\
    CIFAR-100/5 & 5.35 & 1 x NVIDIA A100 40GB \\
    CIFAR-100/10 & 13.53 & 1 x NVIDIA A100 40GB \\
    ImageNet100-64/5 & 30.73 &  4 x NVIDIA A100 40GB \\
    \bottomrule
  \end{tabular}
\end{table}

\section{Forgetting in diffusion models}
\label{app:diff}

\subsection{Coverage of data manifold}
In this analysis, we explore the forgetting behavior of a diffusion model trained continually on CIFAR-10, divided into two tasks with 25000 training samples each. Initially, we train the model exclusively on data from the first task (\textit{Real Task 1}). Subsequently, we employ two approaches: standard self-rehearsal training for the second task (\textit{Continual Task 2}), akin to our method, and retraining the first task's model on the entire CIFAR-10 dataset (\textit{Upper-bound Task 2}), which serves as our upper bound. All training conditions, including architectures, training steps (100K), and hyperparameters, remain consistent across setups. Generative metrics (FID \citep{heusel2017gans}, Precision, and Recall \citep{kynkaanniemi2019improved}) for samples generated with DDIM100 from all models are displayed in Tab.~\ref{tab:coverage_diffusion}.

\begin{table*}[ht]
\caption{Comparison of FID, Recall, and Precision metrics of all three trained models. We calculate metrics for 25000 samples generated using DDIM100. When we train the diffusion model in a continual manner, we observe a very significant drop in the Recall.}
\label{tab:coverage_diffusion}
\centering
\begin{small}
\begin{sc}
\begin{tabular}{ccccccccc}
\toprule
 & FID ($\downarrow$) & Recall ($\uparrow$) &  Precision ($\uparrow$) \\
\toprule
Real Task 1 & 9 & 55 & 65 \\
\midrule
Upper-bound Task 2 & \textbf{13} & \textbf{52} & 64 \\ 
Continual Task 2 & 26 & 37 & \textbf{67} \\
\bottomrule
\end{tabular}
\end{sc}
\end{small}
\end{table*}

\begin{figure}[ht]
    \centering
    \begin{subfigure}[t]{0.3\textwidth}
        \centering
        \raisebox{-\height}{\includegraphics[width=\textwidth]{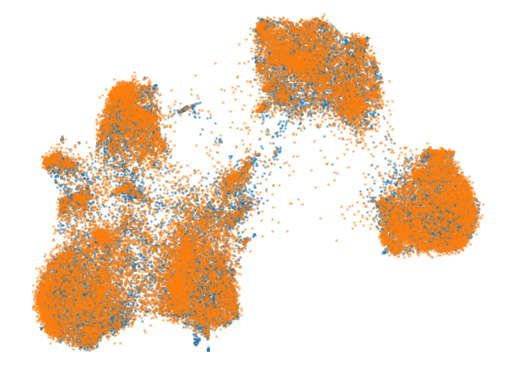}}
        \caption{Initial task 1 data manifold.}
        \label{fig:1}
    \end{subfigure}
    \hfill 
    \begin{subfigure}[t]{0.3\textwidth}
        \centering
        \raisebox{-\height}{\includegraphics[width=\textwidth]{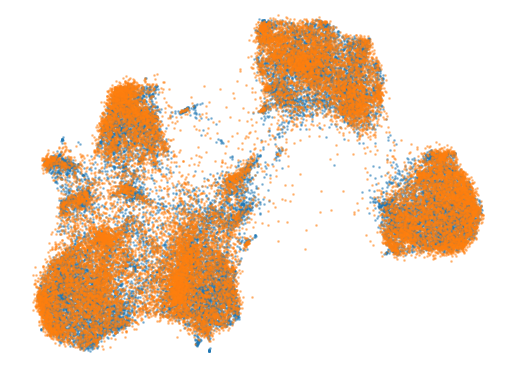}}
        \caption{Task 1 data manifold after training on task 2 -- \textbf{upper-bound}.}
        \label{fig:2}
    \end{subfigure}
    \hfill 
    \begin{subfigure}[t]{0.3\textwidth}
        \centering
        \raisebox{-\height}{\includegraphics[width=\textwidth]{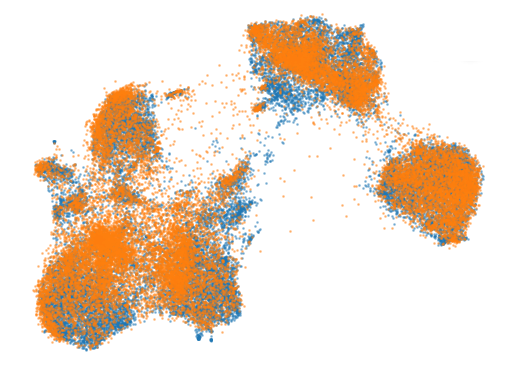}}
        \caption{Task 1 data manifold after training on task 2 -- \textbf{continual training}}
        \label{fig:3}
    \end{subfigure}
    \caption{Visualization of task 1 data manifold in the CIFAR-10 dataset. \textcolor{blue}{Blue points} represent embedded real data samples from task 1, and \textcolor{orange}{orange points} represent generated samples. We see a significant drop in coverage of training data manifold after training on the second task of continual training.}
    \label{fig:three_manifolds}
\end{figure}

Continual training of the diffusion model leads to a noticeable reduction in its capability to cover the training manifold of Task 1, as evidenced by a decrease in the Recall metric. However, the Precision metric does not show a significant drop, indicating that the quality of the generated samples remains largely unaffected. Moreover, we visualize the umap embeddings of both real data examples and generated samples in Fig.~\ref{fig:three_manifolds}. In plot (c), it's evident that the coverage of real data samples (depicted in blue) by the generated samples has noticeably diminished.

\newpage
\subsection{Forgetting of training data samples}

\begin{wrapfigure}{R}{0.4\textwidth}
\vskip -0.2in
\begin{subfigure}[b]{0.4\textwidth}
        \centering
        \includegraphics[width=\textwidth]{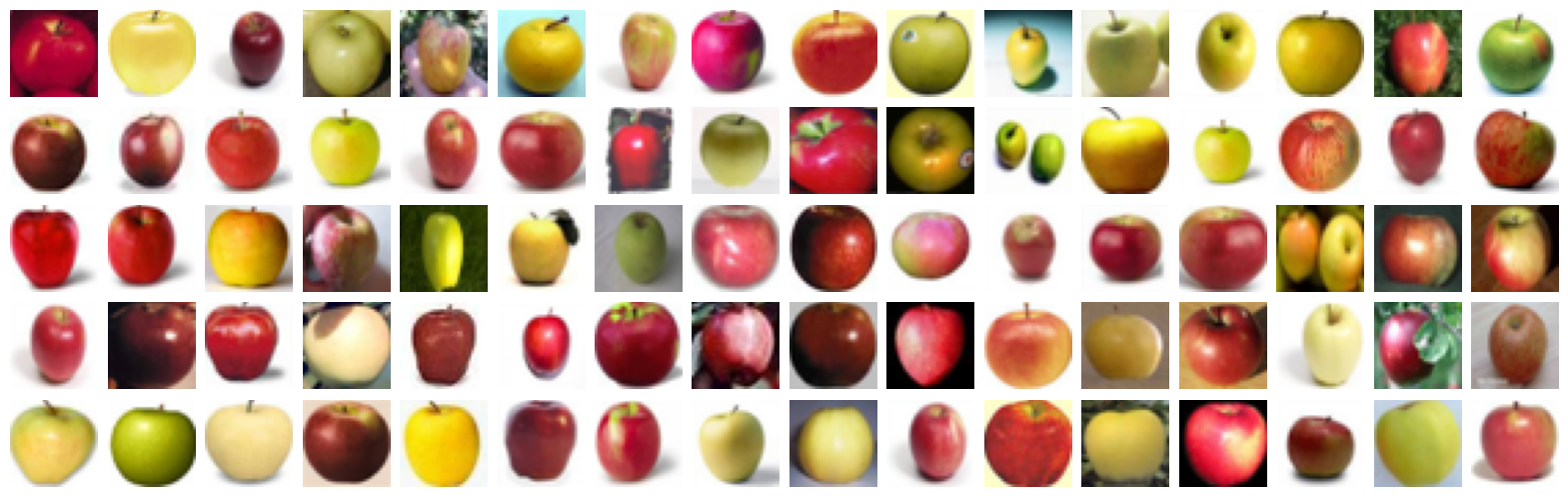}
        \caption{Samples from class \textit{apple} remembered by the diffusion model after training on the second task.}
        \label{fig:remembered_apples}
    \end{subfigure}
    \hfill 
    \begin{subfigure}[b]{0.4\textwidth}
        \centering
        \includegraphics[width=\textwidth]{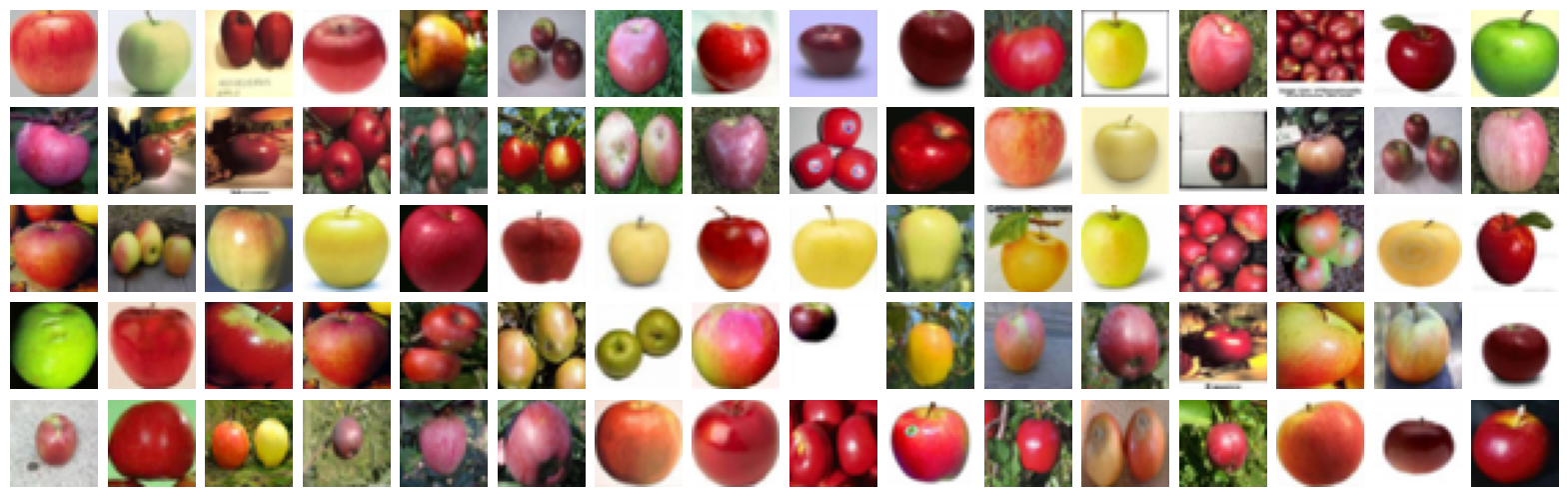}
        \caption{Samples from class \textit{apple} forgotten by the diffusion model after training on the second task.}
        \label{fig:forgotten_apples}
    \end{subfigure}
\end{wrapfigure}

In our study, we further analyze what data samples the diffusion model trained continually tends to forget. We conducted an experiment where we sample examples from class \textit{apple} from our class-conditional diffusion model, trained on the CIFAR-100 dataset divided into 5 tasks, after completing the second task. Utilizing Precision and Recall metrics definitions \citep{kynkaanniemi2019improved}, we computed representations using the Inception-v3 model for both real and generated sample sets. This allowed us to identify which segments of the approximated real data manifold were covered or missed by the generated manifold.

Figure~\ref{fig:remembered_apples} illustrates examples of remembered data samples, while Figure~\ref{fig:forgotten_apples} depicts those data samples that were forgotten by the diffusion model due to continual training. Consistent with findings by \citet{toneva2018empirical}, our diffusion model tends to forget rare samples in the training set, such as those with complex backgrounds, while retaining more common samples characterized by simpler backgrounds and typical shapes. This is also connected to the drop in Recall, where from task to task, the diffusion model loses the ability to generate samples from low-density regions of data manifold \citep{sehwag2022generating}.

\section{Runtime analysis of \ours{} for CIFAR-10/5}
\label{sec:runtime}
We perform an extensive runtime analysis in order to compare the training times of GUIDE to other existing baselines. For the training of each method, we used the same machine with a single NVIDIA RTX A5000 GPU. Since in the training of GUIDE on CIFAR-10/5 setup, we generate new rehearsal examples every 10 batches, for fairness, we evaluate the DDGR runtime with exactly the same interval. We present the calculated runtimes in Tab.~\ref{tab:runtime_analysis}.

\begin{table}[h!]
  \caption{Runtime analysis of all baseline methods on the CIFAR-10/5 benchmark. (*) We evaluate the DDGR method with the same rehearsal generation interval as we use in \ours{} on this benchmark -- once every 10 batches.}
  \label{tab:runtime_analysis}
  \centering
  \begin{tabular}{lc}
    \toprule
    Method & Time [GPU-hours]                  \\
    \midrule
    DGR VAE & 1.28      \\
    DGR+distill   & 1.35    \\
    RTF   & 1.09    \\
    BIR   &  2.49    \\
    GFR   & 1.76    \\
    DDGR & $\text{111.96}^{*}$ \\
    DGR diffusion   & 76.1  \\
    \textbf{GUIDE} & 82.95 \\
    \bottomrule
  \end{tabular}
\end{table}

Approaches that train diffusion models (DDGR, DGR diffusion, and GUIDE) require significantly more computations. However, GUIDE does not increase much of the runtime compared to the standard DGR with diffusion model ($+9\%$ wall time). At the same time, when comparing GUIDE to the recently proposed DDGR \cite{gao2023ddgr} method that also uses diffusion models in generative replay scenario, our approach needs notably less computation time ($-26\%$ wall time) while achieving better performance in all evaluated setups.

\newpage
\section{Speed vs accuracy trade-off in GUIDE}
\label{app:speedup}
We investigate the trade-off between speed and performance of our method. To that end, we study the effect of the number of denoising steps used for rehearsal sampling and the replay's set size on the performance of \ours{}. In both experiments, we present average accuracy after the final task, along with a speed-up in time.

To investigate the effect of the number of denoising steps used for rehearsal sampling, we test the performance of \ours{} for $[10, 20, 50, 100, 250, 1000]$ steps of DDIM (Tab.~\ref{tab:ddim_steps_analysis}). Although the performance tends to increase with a number of denoising steps, the difference in average accuracy is not very significant.  

To effectively reduce the size of the replay set, we generate new rehearsal samples once every $[1,5,10,50,100]$ batches of the training (Tab.~\ref{tab:gen_interval_analysis}). For completeness, we also present the results for fine-tuning setup - training without any generative replay that can be seen as an upper bound in terms of computational complexity. We observe that increasing the generation interval can not only significantly decrease the training time but also increase the final performance.

Although the sampling process of diffusion models is indeed relatively slow compared to other generative architectures, we show that we can significantly reduce the training time of GUIDE without a notable drop in performance by using either one of the proposed techniques.

\begin{table}[h!]
  \caption{Trade-off between the number of denoising steps used for the rehearsal sampling on the effectiveness of GUIDE on CIFAR-10/5 benchmark (default setup from the main text is grayed out).}
  \label{tab:ddim_steps_analysis}
  \centering
  \begin{tabular}{lcc}
    \toprule
    DDIM steps & $\bar{A}_T$ & Time [speedup]  \\
    \midrule
    10     &  $59.7  \pm 1.04$        & $\times$ 7.3\\
    20     &  $63.7  \pm 0.99$     & $\times$ 2.3\\
    \rowcolor{lightergray} 50     & $64.1 \pm 0.09$   & $\times$  1 \\
    100     & $64.2 \pm 0.54$  & $\times$ 0.51 \\
    250     &$63.6 \pm 0.25$ &    $\times$ 0.21 \\
    1000 (no DDIM)     &$71.8 \pm 0.23$ &     $\times$ 0.06 \\
    \bottomrule
  \end{tabular}
\end{table}

\begin{table}[h!]
  \caption{Trade-off between the interval of the generation of new rehearsal samples on the effectiveness of GUIDE on CIFAR-10/5 benchmark (default setup from the main text is grayed out).}
  \label{tab:gen_interval_analysis}
  \centering
  \begin{tabular}{lcc}
    \toprule
    Rehearsal generation interval & $\bar{A}_T$ & Time [speed-up]  \\
    \midrule
    1     & $63.12 \pm 0.45$ &   $\times$  0.2 \\
    \rowcolor{lightergray} 5 & $64.47 \pm 0.45$ & $\times$ 1 \\
    10     & $64.1 \pm 0.09$ &    $\times$ 1.95 \\
    50     & $59.2 \pm 0.47$ &    $\times$ 8.49 \\
    100    & $56.9 \pm 0.98$ &  $\times$ 14.41 \\
    $\infty$ (Fine-tuning) & $18.95 \pm 0.20$   & $\times$ 49.59 \\
    \bottomrule
  \end{tabular}
\end{table}

\vfill 

\section{Samples from ImageNet100 128x128}
\label{app:imagenet_samples}

\begin{figure}[H]
\centering
\includegraphics[width=\columnwidth]{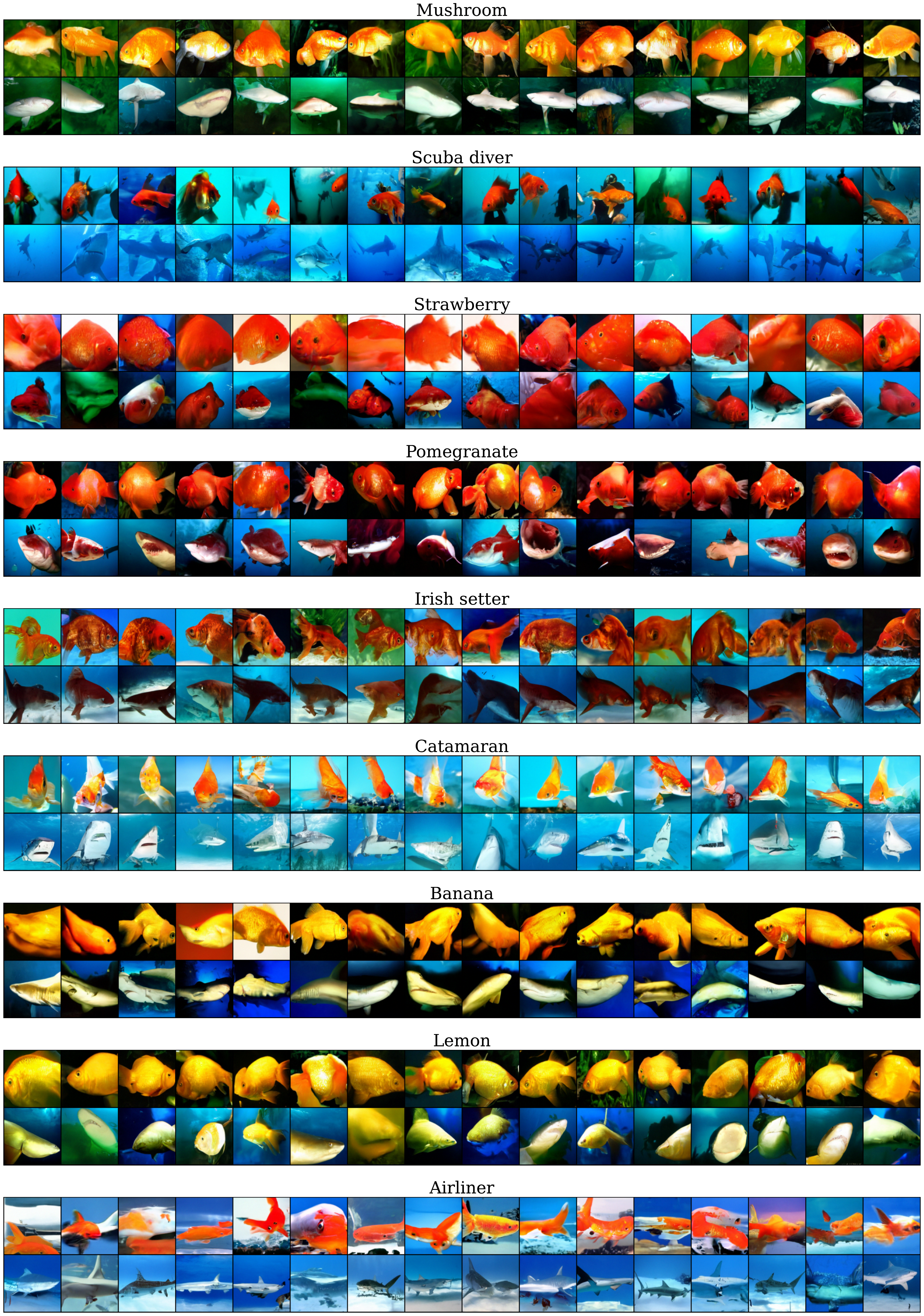}
\end{figure}

\begin{figure}[H]
\centering
\includegraphics[width=\columnwidth]{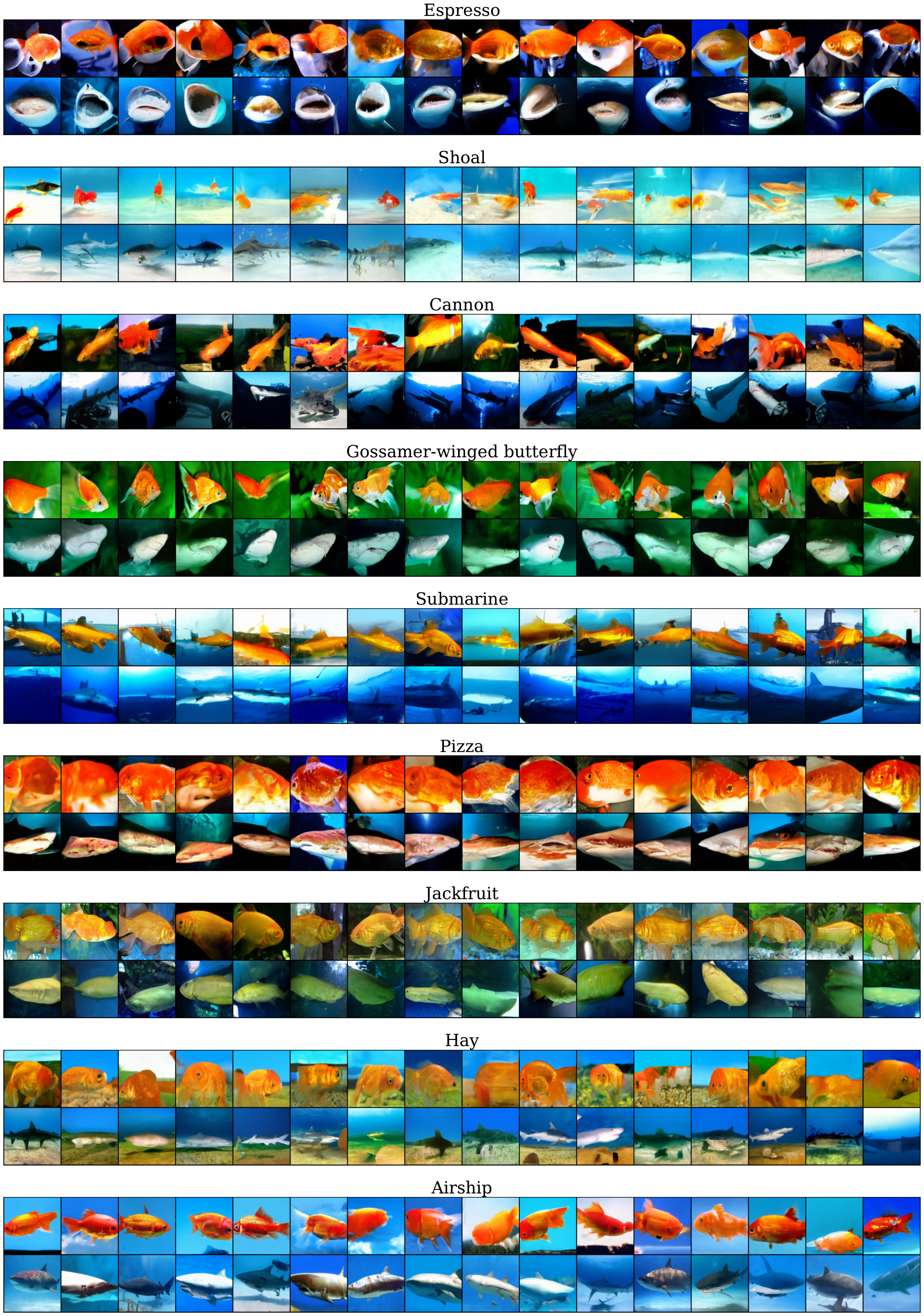}
\end{figure}

\begin{figure}[H]
\centering
\includegraphics[width=\columnwidth]{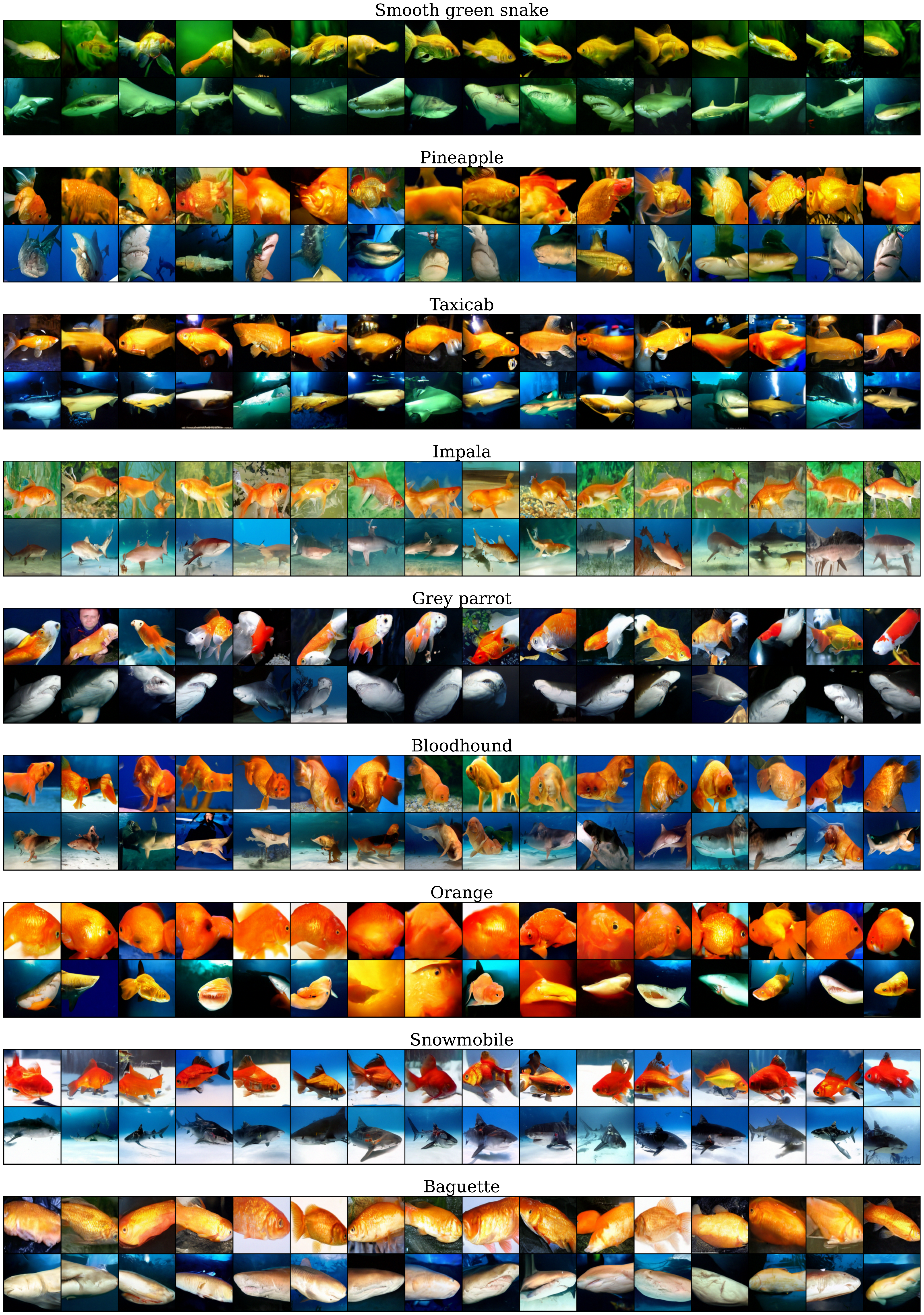}
\caption{Random samples generated from an unconditional diffusion model trained only on \textit{goldfish} and \textit{tiger shark} classes from the ImageNet100 128x128 dataset. Each image grid presents samples generated with guidance to the class depicted above the grid which is unknown to the diffusion model. We generate samples using 1000 denoising steps and we set guidance scale $s$ to $10$ for both classes that we guide to.}
\label{fig:i100_samples1}
\end{figure}

\section{Samples from CIFAR-10}
\begin{figure}[H]
\centering
\includegraphics[width=\columnwidth]{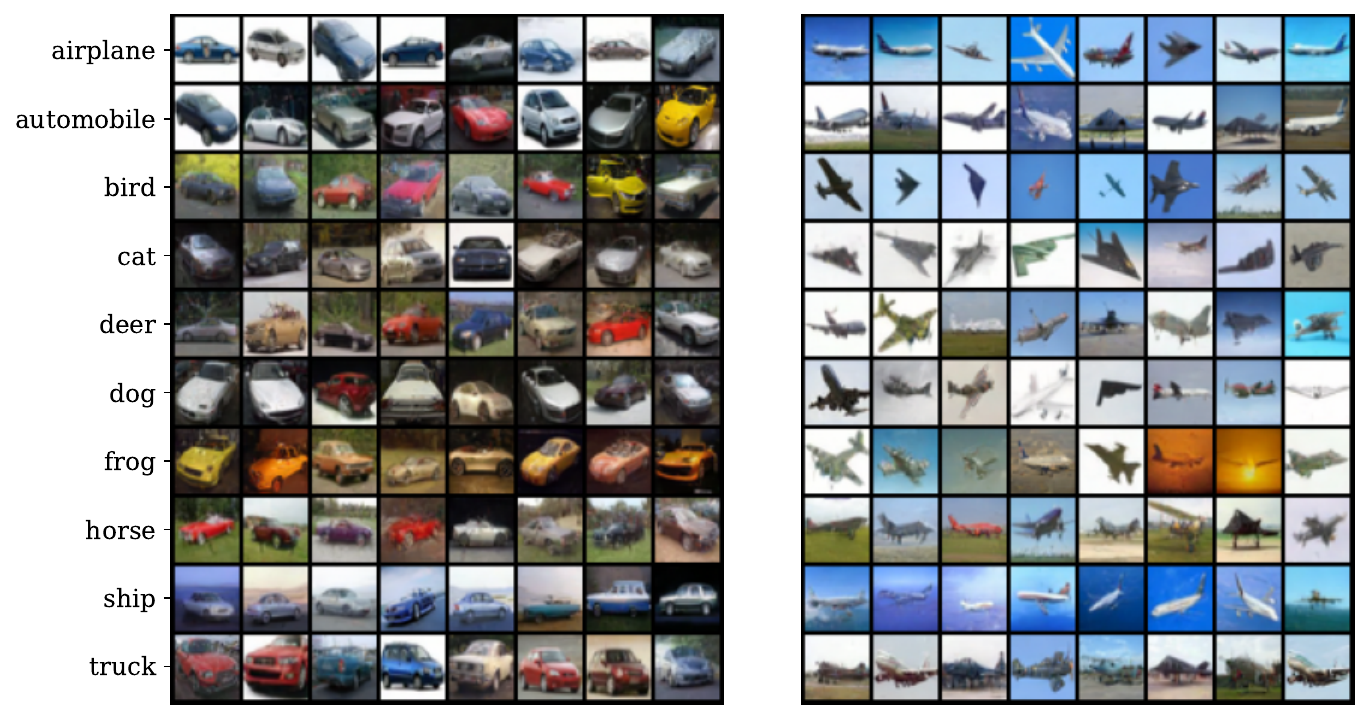}
\caption{Samples from unconditional diffusion model trained only on classes \textit{automobile} and \textit{airplane} from the CIFAR-10 dataset. We generate those samples following the same procedure as discussed in Sec.~\ref{sec:intuition}. On the left grid, we guide the denoising process to automobiles, while on the right grid, to airplanes. At the same time, we add the guidance to classes depicted on the y-axis of the figure. We generate samples using 1000 denoising steps and we set guidance scale $s$ to $5$ for both classes that we guide to.}
\label{fig:cifar10_samples}
\end{figure}

\section{Broader impacts}
\label{sec:broader_impacts}
This paper presents work whose goal is to advance the field of continual machine learning. There are many potential societal consequences of our work, many of which are generic to the machine learning field in general, i.e., the DGR algorithm considered in this paper will reflect the biases present in the dataset. Hence, it is important to exercise caution when using this technique in applications where dataset biases could lead to unfair outcomes for minority and/or under-represented groups. In our case, this especially concerns training a diffusion model used for replay, in which simple random sampling can result in a different data distribution than the original dataset. It would be worthwhile to actively monitor the model's outputs for fairness or implement bias correction techniques to mitigate these negative impacts.

Additionally, while diffusion models have shown promise in various generative tasks, their adaptability through fine-tuning and continual learning remains relatively unexplored. In this early-stage research, some potential risks can emerge while combining our proposed algorithm of guidance with malicious models or when performing more sophisticated attacks.

\end{document}